\pdfoutput=1
\documentclass{article}

\PassOptionsToPackage{numbers, compress}{natbib}

\usepackage[preprint]{neurips_2019}

\usepackage[utf8]{inputenc} 
\usepackage[T1]{fontenc}    
\usepackage{hyperref}       
\usepackage{url}            
\usepackage{booktabs}       
\usepackage{amsfonts}       
\usepackage{nicefrac}       
\usepackage{microtype}      
\usepackage{multirow}
\usepackage{amsmath}
\usepackage{amsthm}
\newtheorem{definition}{Definition}
\DeclareMathOperator*{\argmax}{argmax}
\usepackage{subfig}
\usepackage{wrapfig}
\usepackage{algorithmic} 
\usepackage{algorithm}
\usepackage[pdftex]{graphicx}
\graphicspath{{./fig/}}

\title{Global Adversarial Attacks for Assessing Deep Learning Robustness}

\author{%
  Hanbin Hu \\
  Department of ECE\\
  Texas A\&M University\\
  College Station, TX 77843 \\
  \texttt{hanbinhu@tamu.edu} \\
  \And
  Mit Shah \\
  Department of ECE\\
  Texas A\&M University\\
  College Station, TX 77843 \\
  \texttt{mit09m@tamu.edu} \\
  \And
  Jianhua Z. Huang\\
  Department of Statistics\\
  Texas A\&M University\\
  College Station, TX 77843 \\
  \texttt{jianhua@stat.tamu.edu} \\
  \And
  Peng Li \\
  Department of ECE\\
  Texas A\&M University\\
  College Station, TX 77843 \\
  \texttt{pli@tamu.edu} \\
}

\begin{document}

\maketitle

\begin{abstract}
It has been shown that deep neural networks (DNNs) may be vulnerable to adversarial attacks, raising the concern on their robustness particularly for  safety-critical applications. Recognizing the local nature and limitations of existing adversarial attacks, we present a new type of global adversarial attacks for assessing global DNN robustness. More specifically, we propose a novel concept of global adversarial example pairs in which each pair of two examples are close to each other but  have different class labels predicted by the DNN.  We further propose two families of global attack methods and show that our methods are able to generate diverse and intriguing adversarial example pairs at locations far from the training or testing data. Moreover, we demonstrate that DNNs hardened using the strong projected gradient descent (PGD) based (local) adversarial training are vulnerable to the proposed global adversarial example pairs, suggesting that global robustness must be considered while training robust deep learning networks. 

\end{abstract}

\section{Introduction}
\label{sec:intro}

Deep neural networks (DNNs) have been applied to many applications including safety-critical tasks such as autonomous driving \citep{luckow2016AutoDL} and unmanned autonomous vehicles (UAVs) \citep{carrio2017UAVreview}, which demand high robustness of decision making. However, recently it has been shown that DNNs are susceptible to attacks by adversarial examples \citep{goodfellow2015explaining}. For image classification, for example, adversarial examples may be  generated by adding crafted perturbations indistinguishable to human eyes to legitimate inputs to alter the decision of a trained DNN into an incorrect one.  Several studies attempt to reason about the underlying causes of susceptibility of deep neural networks towards adversarial examples, for instance, ascribing vulnerability  to linearity of the model \citep{goodfellow2015explaining}
or  flatness/curvedness of the decision boundaries \citep{moosavi2017analysis}.  A widely agreed consensus is certainly desirable, which is under ongoing research.

\paragraph{Target global adversarial attack problem.}
The main objectives of this work are to reveal potential vulnerability of DNNs by presenting a new type of attacks, namely \emph{global adversarial attacks}, propose methods for generating such global attacks, and finally demonstrate that DNNs enhanced by conventional (local) adversarial training exhibit little defense to the proposed global adversarial examples. While several adversarial attack methods were proposed \citep{goodfellow2015explaining,papernot2016limitations,carlini2016towards,kurakin2016adversarial,madry2018towards} in recent literature, we refer to these methods as \emph{local adversarial attack methods} as they all aim to solve the \emph{local adversarial attack problem} defined as follows.
\begin{definition}
\textbf{Local adversarial attack problem}. Given an input space $\Omega$, one legitimate input example $\mathbf{x}\in\Omega$ with  label  $y\in\mathcal{Y}$, and a trained DNN $f:\Omega\rightarrow\mathcal{Y}$, find another (adversarial) input example $\mathbf{x}^\prime \in\Omega$  within a radius of $\epsilon$ around $\mathbf{x}$ under a distance measure defined by a norm function $\left\|\cdot\right\|:\left\{\mathbf{x}_a-\mathbf{x}_b\left|\mathbf{x}_a\in\Omega,\mathbf{x}_b\in\Omega\right.\right\}\rightarrow\mathbb{R}_{\geq 0}$ such that $f\left(\mathbf{x}^\prime\right)\neq y, \left\|\mathbf{x}^\prime-\mathbf{x}\right\|\leq\epsilon$.
\end{definition}
Typically, the above problem is solved via optimization governed by a loss function, $\mathcal{L}:\mathcal{Y}\times\mathcal{Y}\rightarrow\mathbb{R}$, measuring the difference between the predicted label $f\left(\mathbf{x}^\prime\right)$ and $y$:
\begin{equation}
\mathbf{x}^{\prime*}=\argmax_{\left\|\mathbf{x}^\prime-\mathbf{x}\right\|\leq\epsilon, \mathbf{x}^\prime\in\Omega}\mathcal{L}\left(f\left(\mathbf{x}^\prime\right),y\right).
\end{equation}
Importantly, the above problem formulation has two key limitations. 
\textbf{(1)} it is deemed \emph{local} in the sense that it only examines  model robustness inside a \emph{local} region centered at each given input $\mathbf{x}$, which in practice is chosen from the training or testing dataset. As such, local adversarial attacks are not adequate since evaluating the DNN robustness around the training and testing data does not provide a complete picture of robustness globally, i.e. in the entire space $\Omega$. On the other hand, assessment of global robustness is essential, e.g. for safety-critical applications. \textbf{(2)} local attack methods assume that for each clean example $\mathbf{x}$ the label $y$ is known. As a result, they are inapplicable to attack the DNN around locations where no labeled data are available. 


In this paper, we  propose a notion of global DNN robustness and a global adversarial attack problem formulation to assess it.  
We evaluate global robustness of a given DNN by assessing the potential high sensitivity of its decision function with respect to small input perturbation leading to change in the predicted label, globally in the entire $\Omega$. By solving the global attack problem, we generate  multiple \emph{global adversarial example pairs} such that each pair of two examples are close to each other but have different labels predicted by the DNN.  The detailed definitions are presented in Section \ref{sec:gadv}.


\paragraph{Related works.} 

Apart from the local adversarial attacks, several other approaches for DNN robustness evaluation have been reported.  
The Lipschitz constant is  utilized to bound DNNs' vulnerability to adversarial attacks \citep{tsuiwei2018clever,yusuke2018lipschitz}. As argued in \citep{ian2018gradmask,todd2018liplimit}, however, currently there is no accurate method for estimating the Lipschitz constant, and the resulting overestimation can easily render its use unpractical. \citep{yang2018gan,isaac2019gan} propose to train a generative model for generating unseen samples for which misclassification happens. However, the ground-truth labels of generated examples must be provided for final assessment and these examples do not capture model's vulnerability due to high sensitivity to small input perturbation.  

\paragraph{Our contributions.}
We propose a new concept of global adversarial examples and several global attack methods. Specifically, we (\textbf{a}) propose a novel concept called global adversarial example pairs and formulate a global adversarial attack problem for assessing the model robustness over the entire input space without extra data labeling; (\textbf{b}) present two families of global adversarial attack methods: (\textbf{1}) alternating gradient adversarial attacks and (\textbf{2}) extreme-value-guided MCMC sampling attack, and demonstrate their effectiveness in generating global adversarial example pairs; (\textbf{c}) using the proposed global attack methods, demonstrate that DNNs hardened using strong projected gradient descent (PGD) based (local) adversarial training are vulnerable towards  the proposed global adversarial example pairs, suggesting that global robustness must be considered while training DNNs.

\section{Global adversarial attacks}
\label{sec:gadv}

We formulate a new global adversarial attack problem as follows.
\begin{definition}
\textbf{Global adversarial attack problem}. Given an input space $\Omega$ and an DNN model $f:\Omega\rightarrow\mathcal{Y}$, find one or more global adversarial example pairs $\left(\mathbf{x}_1, \mathbf{x}_2\right)\in\Omega\times\Omega$  within a radius of $\epsilon$ under a distance measure defined by a norm function $\left\|\cdot\right\|:\left\{\mathbf{x}_a-\mathbf{x}_b\left|\mathbf{x}_a\in\Omega,\mathbf{x}_b\in\Omega\right.\right\}\rightarrow\mathbb{R}_{\geq 0}$ such that $f\left(\mathbf{x}_1\right)\neq f\left(\mathbf{x}_2\right), \left\|\mathbf{x}_1-\mathbf{x}_2\right\|\leq\epsilon$.
\end{definition}
When no confusion occurs, \emph{global adversarial example pair} and \emph{global adversarial examples} are used interchangeably throughout this paper.  
The above problem formulation can be cast into an optimization problem w.r.t. a certain loss function $\mathcal{L}:\mathcal{Y}\times\mathcal{Y}\rightarrow\mathbb{R}$ in the following form:
\begin{equation}
\label{eqn:gadv}
\mathbf{x}_1^*,\mathbf{x}_2^*=\argmax_{\left\|\mathbf{x}_1-\mathbf{x}_2\right\|\leq\epsilon, \left(\mathbf{x}_1,\mathbf{x}_2\right)\in\Omega\times\Omega}\mathcal{L}\left(f\left(\mathbf{x}_1\right),f\left(\mathbf{x}_2\right)\right)
\end{equation}
For convenience of notation, we use $\mathcal{L}_f\left(\mathbf{x}_1,\mathbf{x}_2\right)$ to denote $\mathcal{L}\left(f\left(\mathbf{x}_1\right),f\left(\mathbf{x}_2\right)\right)$.
The above definition and problem formulation have several favorable characteristics. A robust DNN model should be insensitive to small input perturbations. Therefore, two nearby inputs shall share the same (or similar) model output. Conversely, any large $\mathcal{L}_f\left(\mathbf{x}_1,\mathbf{x}_2\right)$ value of two nearby inputs reveals a sharp transition over the decision boundary in a classification task or an unstable region in a regression task.

\begin{wrapfigure}{r}{0.4\linewidth}
\centering
\includegraphics[width=\linewidth]{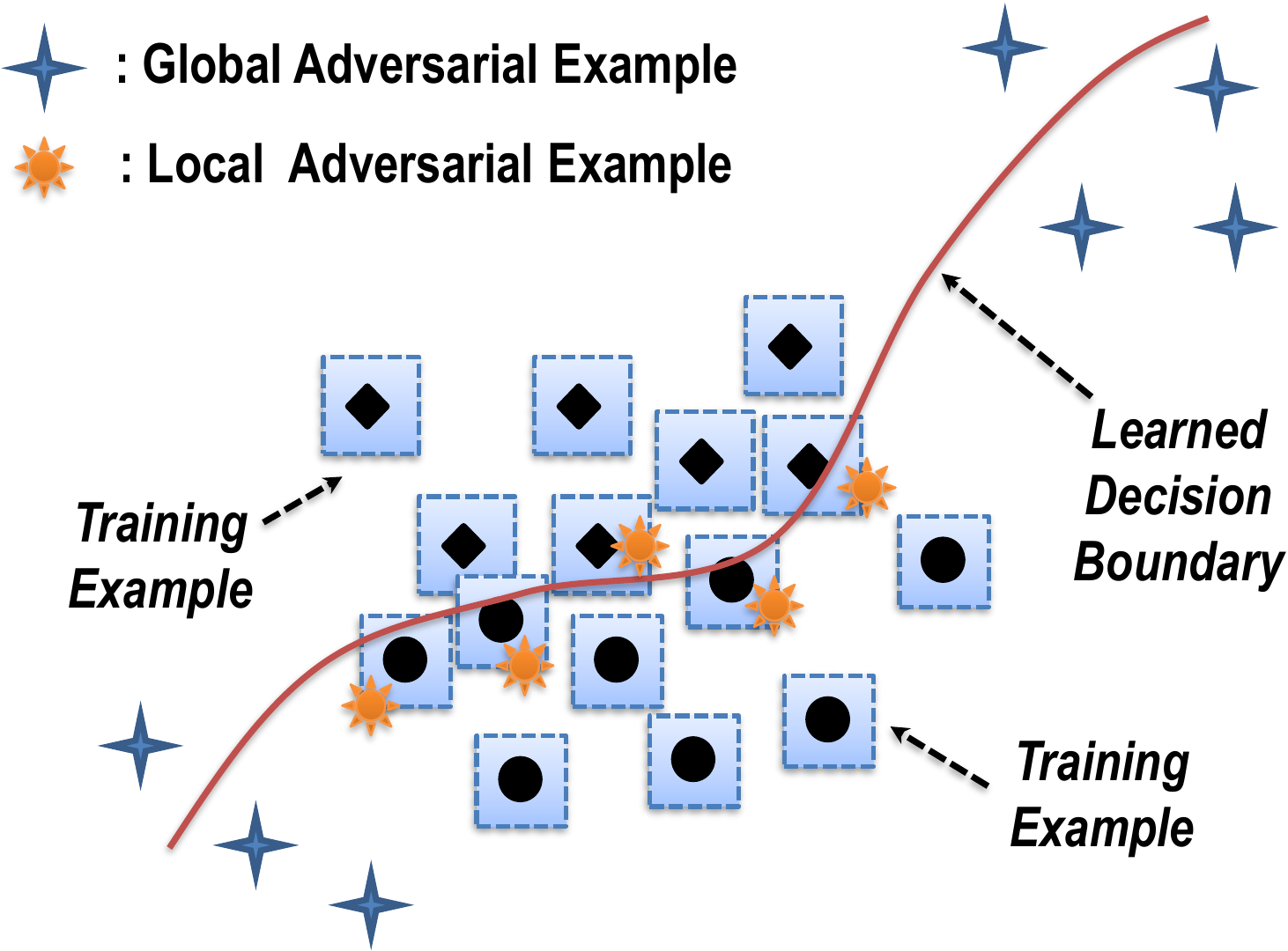}
\caption{Global vs. local adversarial examples.}
\label{fig:gadv}
\end{wrapfigure}

Loosely speaking, the maximum of $\mathcal{L}_f\left(\mathbf{x}_1,\mathbf{x}_2\right)$ over the entire input space can serve as a measure of global model robustness. If it is larger than a preset threshold, the DNN may be deemed as vulnerable towards global adversarial attacks. On the other hand,  (\ref{eqn:gadv}) is a global optimization problem for which multiple near-optimal solutions may be reached by starting from different initial solutions or employing different optimization methods. Practically, any pair of two close inputs $\mathbf{x}_1,\mathbf{x}_2$ with different model predictions in the case of classification or a sufficiently large value of   $\mathcal{L}_f\left(\mathbf{x}_1,\mathbf{x}_2\right)$ in the case of regression may be considered as a global adversarial example pair. 

It is important to note that our problem formulation doesn't restrict  adversarial examples  to be around a certain input example as in the case of the existing local adversarial attacks; it only sets an indistinguishable distance between a pair of two input examples in order to examine the entire input space, e.g. at locations far away from the training or testing dataset. Fig. \ref{fig:gadv} contrasts the conventional local adversarial examples with the proposed global adversarial examples. Importantly, our global attack formulation does not require additional labeled data;  it directly measures the model's sensitivity to input perturbation  and be applied globally in the entirety of the input space. 

We propose two families of attack methods to solve (\ref{eqn:gadv}) as a way of generating global adversarial examples:  \textbf{1}) alternating gradient global adversarial attacks and \textbf{2}) extreme-value-guided MCMC sampling global adversarial attack, as discussed in Section \ref{sec:gbatk} and Section \ref{sec:sampleatk}, respectively.

\section{Alternating gradient global adversarial attacks}
\label{sec:gbatk}

\begin{wrapfigure}{r}{0.18\linewidth}
\centering
\includegraphics[width=\linewidth]{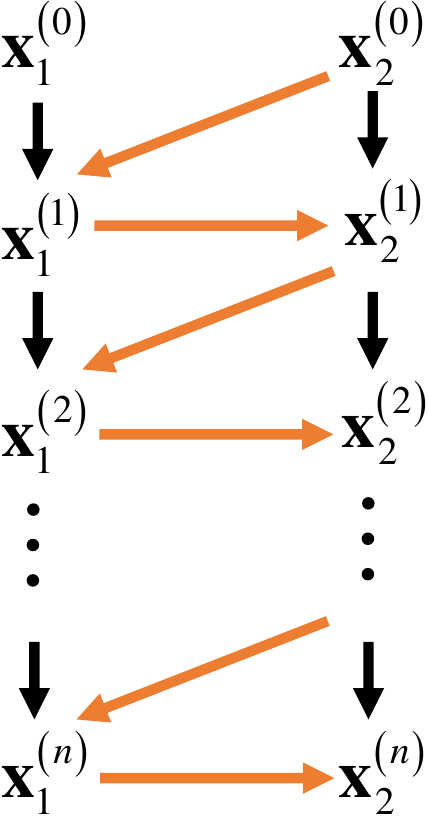}
\caption{Alternating attack illustration.}
\label{fig:gbatk}
\end{wrapfigure}

For global adversarial example pair generation per (\ref{eqn:gadv}), a pair of two examples $\left(\mathbf{x}_1,\mathbf{x}_2\right)$ shall be be optimized to maximize the loss  $\mathcal{L}_f\left(\mathbf{x}_1,\mathbf{x}_2\right)$ under a distance constraint. We propose a family of attack methods called alternating gradient global adversarial attacks which proceed as follows: 1) start from an initial input pair;  2) fix the first example, and then attack (move) the second under the distance constraint while maximizing the loss $\mathcal{L}_f\left(\mathbf{x}_1,\mathbf{x}_2\right)$  using a gradient-based local adversarial attack method, referred to as a \emph{sub-attack method} here, 3) swap the roles of the first and (updated) second examples, i.e. fix the second example while attacking the first, 4) repeat this process for a number of iterations, as shown in Fig. \ref{fig:gbatk}.  

Given a DNN model $f$ and a loss function $\mathcal{L}$, a sub-attack method can be characterized using a function $\mathbf{x}^\prime=f_{s\_attk}\left(\mathbf{x},y,\mathcal{R}_{\mathbf{x},\epsilon};f,\mathcal{L}\right)$ constructing an adversarial example $\mathbf{x}^\prime$ w.r.t  example $\mathbf{x}$ and its corresponding label $y$, where $\mathcal{R}_{\mathbf{x},\epsilon}=\left\{\mathbf{x}+\delta|\left\|\delta\right\|\leq\epsilon\right\}$ specifies the region for adversarial sample generation. Suppose we start with an initial example pair $\left(\mathbf{x}_1^{\left(0\right)},\mathbf{x}_2^{\left(0\right)}\right)$, then we  get the $\left(i+1\right)$-round example pair from the $\left(i\right)$-round sample pair by:
\begin{eqnarray}
\mathbf{x}_1^{\left(i+1\right)}&=&f_{s\_attk}\left(\mathbf{x}_1^{\left(i\right)},f\left(\mathbf{x}_2^{\left(i\right)}\right),\mathcal{R}_{\mathbf{x}_2^{\left(i\right)},\epsilon};f,\mathcal{L}\right)\label{eqn:ggb1}\\
\mathbf{x}_2^{\left(i+1\right)}&=&f_{s\_attk}\left(\mathbf{x}_2^{\left(i\right)},f\left(\mathbf{x}_1^{\left(i+1\right)}\right),\mathcal{R}_{\mathbf{x}_1^{\left(i+1\right)},\epsilon};f,\mathcal{L}\right)\label{eqn:ggb2}
\end{eqnarray}

Here, to attack one example, we use the other example's model prediction as the label when applying the sub-attack method $f_{s\_attk}$ so that the difference between two examples' model predictions is maximized. In the meanwhile, the search region for attacking one example is  constrained (centered) by the other example. Hence, the distance between the two examples is always less than $\epsilon$. Equations \ref{eqn:ggb1} and \ref{eqn:ggb2} are invoked to alternately attack $\mathbf{x}_1^{\left(i\right)}$ and $\mathbf{x}_2^{\left(i\right)}$ to generate an updated example pair for the next round. As the global attack continues, multiple global adversarial sample pairs may be generated along the way.

\begin{algorithm}
\caption{Global PGD attack algorithm for classification.}
\label{algo:G-PGD}
\begin{algorithmic}[1]
\REQUIRE ~~ \\
Initial starting example pair $\left(\mathbf{x}_1^{\left(0\right)},\mathbf{x}_2^{\left(0\right)}\right)$; Example vector dimension $D$; distance constraint $\epsilon$;
Number of total rounds $N$; Number of sub-attack steps $S$; Step size $a$ for the sub-attack.\\
\ENSURE ~~ 
The set of generated global adversarial sample pairs $T$;
\STATE $T=\left\{\right\}$
\FOR{$i\leftarrow1$ to $N$}
\STATE Sample a uniform distribution perturbation $\delta\sim \mathrm{U}\left[-\epsilon,\epsilon\right]^D$ \label{algo:gpgd:noise1}
\STATE $\mathbf{x}_1^{\left(i\right)}\leftarrow Clip\left(\mathbf{x}_1^{\left(i-1\right)}+\delta,\mathcal{R}_{\mathbf{x}_2^{\left(i-1\right)},\epsilon}\right)$
\FOR{$j\leftarrow1$ to $S$}
\STATE $\mathbf{x}_1^{\left(i\right)}\leftarrow Clip\left(\mathbf{x}_1^{\left(i\right)}+a\cdot\mathrm{sign}\left(\nabla_{\mathbf{x}_1}\mathcal{L}_f\left(\mathbf{x}_1^{\left(i\right)},\mathbf{x}_2^{\left(i-1\right)}\right)\right),\mathcal{R}_{\mathbf{x}_2^{\left(i-1\right)},\epsilon}\right)$
\ENDFOR
\STATE Sample a uniform distribution perturbation $\delta\sim \mathrm{U}\left[-\epsilon,\epsilon\right]^D$ \label{algo:gpgd:noise2}
\STATE $\mathbf{x}_2^{\left(i\right)}\leftarrow Clip\left(\mathbf{x}_2^{\left(i-1\right)}+\delta,\mathcal{R}_{\mathbf{x}_1^{\left(i\right)},\epsilon}\right)$
\FOR{$j\leftarrow1$ to $S$}
\STATE $\mathbf{x}_2^{\left(i\right)}\leftarrow Clip\left(\mathbf{x}_2^{\left(i\right)}+a\cdot\mathrm{sign}\left(\nabla_{\mathbf{x}_2}\mathcal{L}_f\left(\mathbf{x}_2^{\left(i\right)},\mathbf{x}_1^{\left(i\right)}\right)\right),\mathcal{R}_{\mathbf{x}_1^{\left(i\right)},\epsilon}\right)$
\ENDFOR
\STATE $T\leftarrow T\cup \left\{\left(\mathbf{x}_1^{\left(i\right)},\mathbf{x}_2^{\left(i\right)}\right)\right\}$
\ENDFOR
\end{algorithmic}
\end{algorithm}

\paragraph{Choice of sub-attack methods.} We leverage the popular gradient-based local adversarial attack methods as sub-attack methods under the above family of global attacks.   Particularly, the fast gradient sign method (FGSM) \citep{goodfellow2015explaining}, iterative FGSM (IFGSM) \citep{kurakin2016adversarial}, and projected gradient descent (PGD) \citep{madry2018towards} may be considered for sub-attack method $f_{s\_attk}$. As one example, we present the algorithm flow for the global PGD (G-PGD) attack targeting classifiers with PGD employed as the sub-attack in Algorithm \ref{algo:G-PGD}, where the $Clip\left(\mathbf{x},mathcal{R}\right)$ function drags the input $\mathbf{x}$ outside the preset region $\mathcal{R}$ onto the boundary of $\mathcal{R}$. The global IFGSM (G-IFGSM) skips the random noise perturbations (steps \ref{algo:gpgd:noise1} and \ref{algo:gpgd:noise2}). The global FGSM (G-FGSM) sets the number of sub-attack steps $S$ to be $1$ and the step size for sub-attack $a$ to be $\epsilon$ in addition to ignoring the random noise perturbation steps.

\section{Extreme-value-guided MCMC sampling attack}
\label{sec:sampleatk}

While the family of alternating gradient global adversarial attacks discussed in Section~\ref{sec:gbatk} can work effectively in practice, such methods may get trapped at a local maximum, degrading the quality of global attack. To this end, we propose a stochastic optimization approach based on the extreme value distribution theory and Markov Chain Monte Carlo (MCMC) method, which is more advantageous from a global optimization point of view.

\paragraph{Extreme value distribution.}
Consider sampling a set of i.i.d input example pairs $\left\{\left(\mathbf{x}_{1,1},\mathbf{x}_{2,1}\right),\cdots,\left(\mathbf{x}_{1,n},\mathbf{x}_{2,n}\right)\right\}$ in the input pair space $\Omega\times\Omega$. The greatest loss value $L^*=\max_{i\in\left[1,n\right]} \mathcal{L}_f\left(\mathbf{x}_{1,i},\mathbf{x}_{2,i}\right)$ can  be regarded as a random variable following a certain distribution characterized by its density function $p_{L^*}\left(l\right)$. The Fisher-Tippett-Gnedenko theorem says that the distribution of the maximum value of examples, if exists, can only be one of the three families  of extreme value distribution: the Gumbel class, the Fr{\'e}chet and the reverse Weibull class \citep{gomes2015extreme}. Hence, $p_{L^*}\left(l\right)$ falls into one of the three families as well, whose cumulative density function (CDF) $F_{L^*}\left(l\right)$ can be written in a unified form, called the generalized extreme value (GEV) distribution:
\begin{equation}
F_{L^*}\left(l\right)=\left\{
\begin{array}{ll}
\exp\left(-\left(1+\xi\frac{l-\mu}{\sigma}\right)^{-1/\xi}\right) & \xi\neq 0 \\
\exp\left(-\exp\left(-\frac{l-\mu}{\sigma}\right)\right)    & \xi= 0
\end{array}
\right.
\end{equation}
where $\mu$, $\sigma$ and $\xi$ are the location, scale, and shape parameter for the GEV distribution and may be obtained through the maximum likelihood estimation (MLE).

Assuming that the desired  generalized extreme value (GEV) distribution of the loss function is available, multiple large loss values,  corresponding to potential global adversarial example pairs, may be generated by sampling the GEV distribution. The added benefit here is that the inherent randomness in this sampling process may be explored to find more globally optimal solutions. 
Nevertheless, since the GEV distribution may not be easily sampled directly, We adopt  the  Markov Chain Monte Carlo (MCMC) method to sample the GEV distribution \citep{andrieu2003mcmc}. In reality, the GEV distribution is not known \emph{a priori} and shall be estimated using  MLE based on a sample of data as described below.  

\paragraph{Extreme-value-guided MCMC sampling algorithm (GEVMCMC).}
The proposed GEVMCMC algorithm is shown in Algorithm \ref{algo:GEVMCMC} for the case of classification problems. 
There exist two essential components for MCMC sampling: the target distribution, which is in this case  the desired GEV distribution, and the proposal distribution, which is a surrogate distribution easy to sample.  For each MCMC round, we collect an example from the proposal distribution, and then accept this example or discard it while keeping the previous one based on an acceptance ratio $p_A$ in Step \ref{algo:GEVMCMC:pa} of Algorithm \ref{algo:GEVMCMC} \citep{andrieu2003mcmc}. 
In order to sample the block maximum for the extreme value distribution, a block of example pairs are collected from the proposal distribution in each round as in Step \ref{algo:GEVMCMC:block} of the algorithm instead of a single example.
As the MCMC sampling process proceeds, the actual sampling distribution implemented converges to the target (GEV) distribution. 

\begin{algorithm}
\caption{Extreme-value-guided MCMC sampling algorithm (GEVMCMC) for global attack.}
\label{algo:GEVMCMC}
\begin{algorithmic}[1]
\REQUIRE ~~
Initial starting example pair $\left(\mathbf{x}_1^{\left(0\right)},\mathbf{x}_2^{\left(0\right)}\right)$; Number of total rounds $N$; Number of warm-up rounds $N_w$; Block size $B$; Number of example pairs $k$ used for GEV distribution update;\\
\ENSURE ~~
The set of the generated global adversarial example pairs $T$;
\STATE Apply G-PGD for $N_w$ rounds to warm-up and update T. \label{algo:GEVMCMC:warmup}
\FOR{$i\leftarrow N_w+1$ to $N$}
\STATE Sample $B$ i.i.d. samples $\left\{\left(\mathbf{x}_1^{\left[1\right]},\mathbf{x}_2^{\left[1\right]}\right),\cdots,\left(\mathbf{x}_1^{\left[B\right]},\mathbf{x}_2^{\left[B\right]}\right)\right\}$ from the proposal distribution \label{algo:GEVMCMC:block} $q\left(\mathbf{x}_1,\mathbf{x}_2|\mathbf{x}_1^{\left(i-1\right)},\mathbf{x}_2^{\left(i-1\right)}\right)$
\STATE $\left(\mathbf{x}_1^*,\mathbf{x}_2^*\right)\leftarrow\argmax_{j\in\left[1,B\right]}\mathcal{L}_f\left(\mathbf{x}_1^{\left[j\right]},\mathbf{x}_2^{\left[j\right]}\right)$
\STATE Update the GEV distribution $p_{L^*}\left(l\right)$ using top $k$ loss values in the history. \label{algo:GEVMCMC:EVupdate}
\STATE $p_A\leftarrow\min\left\{1.0,\frac{p_{L^*}\left(\mathcal{L}_f\left(\mathbf{x}_1^*,\mathbf{x}_2^*\right)\right)q\left(\mathbf{x}_1^{\left(i-1\right)},\mathbf{x}_2^{\left(i-1\right)}|\mathbf{x}_1^*,\mathbf{x}_2^*\right)}{p_{L^*}\left(\mathcal{L}_f\left(\mathbf{x}_1^{\left(i-1\right)},\mathbf{x}_2^{\left(i-1\right)}\right)\right)q\left(\mathbf{x}_1^*,\mathbf{x}_2^*|\mathbf{x}_1^{\left(i-1\right)},\mathbf{x}_2^{\left(i-1\right)}\right)}\right\}$ \label{algo:GEVMCMC:pa}
\STATE Sample a uniform random variable $\alpha\sim\mathrm{U}\left[0,1\right]$
\IF{$\alpha \leq p_A$}
\STATE Accept the new example. $\left(\mathbf{x}_1^{\left(i\right)},\mathbf{x}_2^{\left(i\right)}\right)=\left(\mathbf{x}_1^*,\mathbf{x}_2^*\right)$
\ELSE
\STATE Reject and keep the previous example. $\left(\mathbf{x}_1^{\left(i\right)},\mathbf{x}_2^{\left(i\right)}\right)=\left(\mathbf{x}_1^{\left(i-1\right)},\mathbf{x}_2^{\left(i-1\right)}\right)$
\ENDIF
\STATE $T\leftarrow T\cup \left\{\left(\mathbf{x}_1^{\left(i\right)},\mathbf{x}_2^{\left(i\right)}\right)\right\}$
\ENDFOR
\end{algorithmic}
\end{algorithm}


Importantly, the generalized extreme value (GEV) distribution $p_{L^*}\left(l\right)$ of the loss function  doesn't exist at the beginning of the algorithm. Therefore,  it is estimated during the sampling process. Two techniques are considered to obtain an accurate GEV distribution. A warm-up procedure (Step \ref{algo:GEVMCMC:warmup} in Algorithm \ref{algo:GEVMCMC}) using a few rounds of G-PGD is performed first to collect a few global adversarial example pairs of large loss values. In each round, the top k loss values among all  example pairs in the history including the ones in the current block are selected  to   estimate $p_{L^*}\left(l\right)$ based on MLE in Step \ref{algo:GEVMCMC:EVupdate} of Algorithm \ref{algo:GEVMCMC}.

\paragraph{Proposal distribution design.}

The convergence speed of MCMC sampling towards the target distribution critically  depends on the proposal distribution~\citep{andrieu2003mcmc}. To efficiently generate  high-quality  global adversarial example pairs, we consider the following essential aspects in designing the proposal distribution:  (\textbf{1}) finding large loss values; (\textbf{2}) enabling global search; (\textbf{3}) constraining two examples in each pair to be within distance $\epsilon$. 

\begin{wrapfigure}{r}{0.25\linewidth}
\centering
\includegraphics[width=\linewidth]{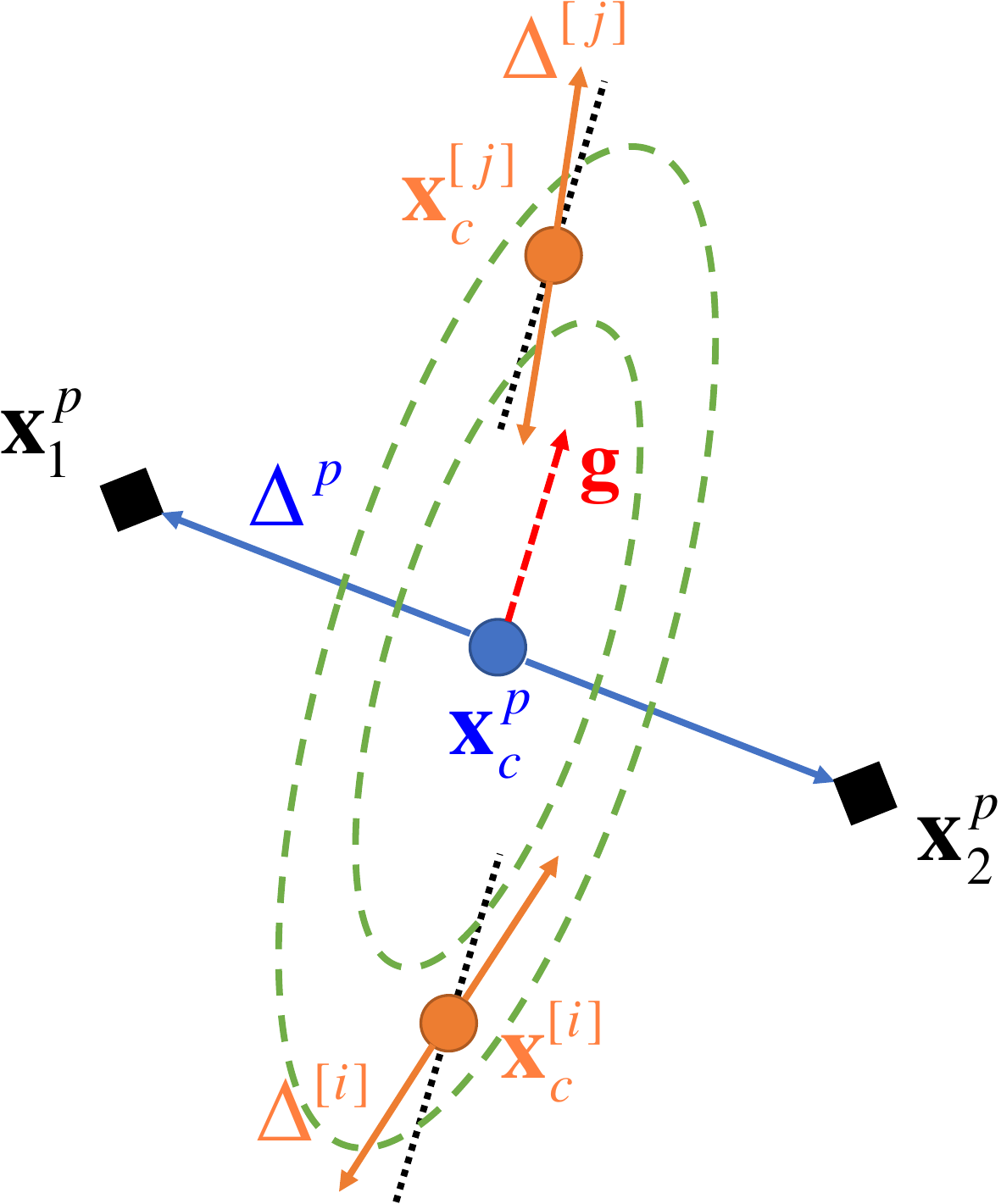}
\caption{MCMC proposal distribution design illustration.}
\label{fig:mcmcpropdist}
\end{wrapfigure}

We decompose the sampling of an example pair into two sequential steps:  sample the center location  $\mathbf{x}_c$ and sample the difference vector $\Delta$ between the two examples. Then, we construct the example pair by $\mathbf{x}_1=\mathbf{x}_c+\Delta, \mathbf{x}_2=\mathbf{x}_c-\Delta$, as shown in Fig.~\ref{fig:mcmcpropdist}. The two sampling steps are independent of each other, and hence, the proposal   distribution conditioning on the previous example pair $\left(\mathbf{x}_c^p,\Delta^p\right)$ is split into a product of a center proposal and difference proposal distribution:
\begin{equation}
q\left(\mathbf{x}_c,\Delta|\mathbf{x}_c^p,\Delta^p\right)=q_{x^c}\left(\mathbf{x}_c|\mathbf{x}_c^p,\Delta^p\right)q_{\Delta}\left(\Delta|\mathbf{x}_c^p,\Delta^p\right).
\end{equation}
As such, the distance constraint for the two examples is only taken into consideration in the design of $q_{\Delta}\left(\Delta|\mathbf{x}_c^p,\Delta^p\right)$ and no distance constraint is necessary when sampling the center.  

We speed up the convergence of MCMC by incorporating  the (normalized) gradient information
\begin{equation}
\mathbf{g}=\frac{\nabla_{\mathbf{x}_c}\mathcal{L}_f\left(\mathbf{x}_c^p+\Delta^p,\mathbf{x}_c^p-\Delta^p\right)}{\left|\nabla_{\mathbf{x}_c}\mathcal{L}_f\left(\mathbf{x}_c^p+\Delta^p,\mathbf{x}_c^p-\Delta^p\right)\right|}
\end{equation}
into the proposal distribution design. We design the center proposal distribution to be a multi-variate Gaussian distribution centered at $\mathbf{x}_c^p$ with a covariance matrix biasing to  sampling along the gradient direction $\mathbf{g}$ to increase the likelihood of finding large loss values while allowing sampling in other directions during the same time:
\begin{equation}
q_{x^c}\left(\mathbf{x}_c|\mathbf{x}_c^p,\Delta^p\right)=\mathcal{N}\left(\mathbf{x}_c^p,\lambda_0^2\mathbf{I}+\left(\lambda_m^2-\lambda_0^2\right)\mathbf{g}\mathbf{g}^{\mathrm{T}}\right),
\end{equation}
where $\lambda_m^2$ sets the largest eigenvalue and $\lambda_0^2 < \lambda_m^2$ sets all other eigenvalues of the covariance matrix. The absolute values of $\lambda_0$ and $\lambda_m$ control the size of the search region while  the ratio between $\lambda_m$ and $\lambda_0$ determines to what extent we want to focus on the gradient direction. 

The gradient sign information is incorporated into the difference proposal distribution considering the distance constraint $\epsilon$. 
Particularly, for the  $l_\infty$ norm based distance measure, we propose a Bernoulli distribution with parameter $p_B> 0.5$ for each element of the difference vector:
\begin{equation}
\Delta_i=\left\{\begin{array}{ll}
    \frac{\epsilon}{2}\mathrm{sign}\left(g_i\right) & \textrm{with probability $p_B$} \\
    -\frac{\epsilon}{2}\mathrm{sign}\left(g_i\right) & \textrm{with probability $1-p_B$},
\end{array}\right.
\end{equation}
which ensures that the pair $\left(\mathbf{x}_1,\mathbf{x}_2\right)$  be within distance $\epsilon$ and each difference component is more likely to be set according to the corresponding gradient sign component. 

\section{Experimental results}
\label{sec:res}

\paragraph{Experimental settings.}

We  investigate several local and global methods on two popular image classification datasets: MNIST \citep{lecun1998gradient} and  CIFAR10 \citep{krizhevsky2009learning}. 
To evaluate DNN robustness globally in the input space, we create an additional class ``meaningless'' and append 6,000 and 5,000 random noisy images (one tenth of the original training dataset) under this ``meaningless'' class into the original MNIST and CIFAR10 training datasets, respectively, and refer to the expanded training datasets as the augmented  training datasets. All trained DNNs perform classification across 11 classes. 

For MNIST, we train a neural network with two convolutional and two fully-connected layers with an accuracy of $99.43\%$ with 40 training epochs. For CIFAR10, a VGG16 \citep{simonyan2014vgg} network is trained for 300 epochs, reaching  $94.25\%$ accuracy. Furthermore, we globally attack  adversarially-trained models,  which are trained using adversarial training  based on local adversarial examples. In each epoch of adversarial training, the adversarial samples are generated by attacking the DNN model from the last epoch using a 30-step local white-box PGD attack, which is considered a strong first-order attack \citep{madry2018towards}.   And then an updated model is trained using both the augmented training set and the generated adversarial images. Adversarial training process is performed for additional 40 epochs for the MNIST model and 30 epochs for the CIFAR10 model, respectively. The ratio of the weighting parameters between the losses of the examples in the augmented training set and adversarial examples are $1:1$.

\paragraph{Adversarial attack parameter settings.}

The $l_{\infty}$ norm based perturbation limit (the maximum allowed difference between two close images) is set to $\epsilon_{MNIST} = 0.1$ for MNIST and to $\epsilon_{CIFAR10}=0.005$ for  CIFAR10. We add another 10,000 random images with the ``meaningless'' class label into the original testing dataset (10,000 samples) to create an augmented  testing dataset. 
We experiment  three common local adversarial attack methods: FGSM \citep{goodfellow2015explaining}, IFGSM \citep{kurakin2016adversarial}, and PGD \citep{madry2018towards}, referred to as L-FGSM, L-IFGSM and L-PGD in this paper. Both L-IFGSM and L-PGD perform a 30-step attack with a $l_{\infty}$ step size of $\epsilon/10$.
All local attacks are performed on the augmented testing set.

All four proposed global adversarial attack methods are considered: alternating gradient global adversarial attack with different sub-attack methods of FGSM (G-FGSM), IFGSM(G-IFGSM) and PGD (G-PGD); extreme-value-guided sampling global attack (GEVMCMC). 
For all global attack methods, we randomly pick 100 images from the original testing dataset and from the appended random testing dataset, respectively, to form the first images of the 100 starting pairs. The second image of each pair is obtained by adding small uniformly-distributed random noise bounded by perturbation size $\epsilon$ to the first image. 100 rounds of optimization are performed by all global adversarial attack methods, generating a two sets of 10,000 adversarial example pairs, one set for each of the two starting conditions: start with the 100 original testing images, start from the 100 appended random testing images.  
G-IFGSM and G-PGD share the same parameter settings with their local attack counterparts  L-IFGSM and L-PGD, respectively.  The number of GEVMCMC initial G-PGD warm-up rounds  is 10 for MNIST and 30 for CIFAR10. The block size $B$ is set to be 59. Three parameters for the proposal distribution for MNIST are $\lambda_m=1.2\epsilon_{MNIST},\lambda_0=0.3\epsilon_{MNIST}, p_B=0.95$, and for CIFAR10 they are set to be $\lambda_m=4.8\epsilon_{CIFAR10},\lambda_0=0.6\epsilon_{CIFAR10}, p_B=0.99$.

\begin{figure}
\centering
\subfloat[\label{sfig::CIFAR10_example:a}]{
\includegraphics[width=0.3\linewidth]{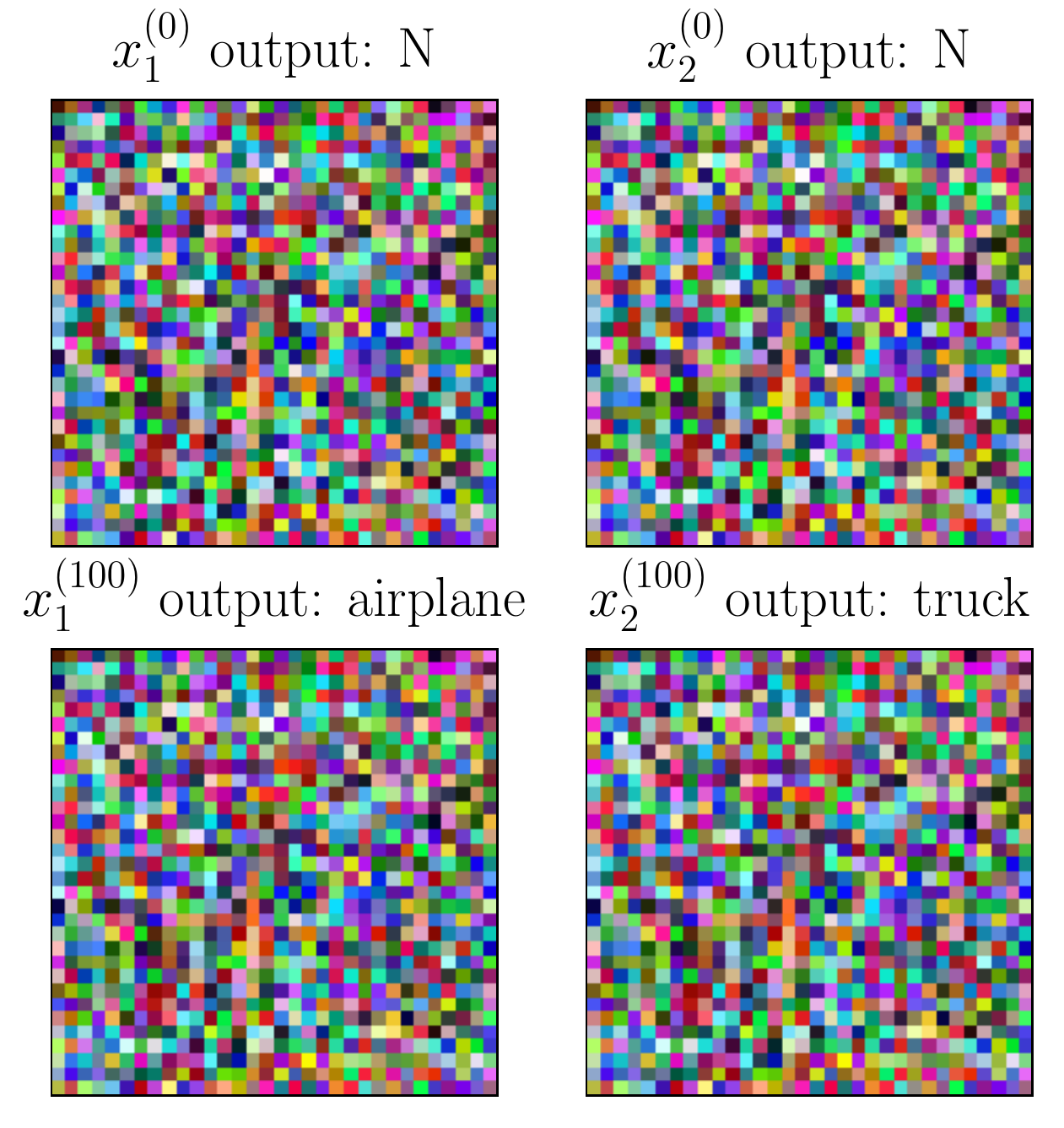}
}
\hfill
\subfloat[\label{sfig::CIFAR10_example:b}]{
\includegraphics[width=0.31\linewidth]{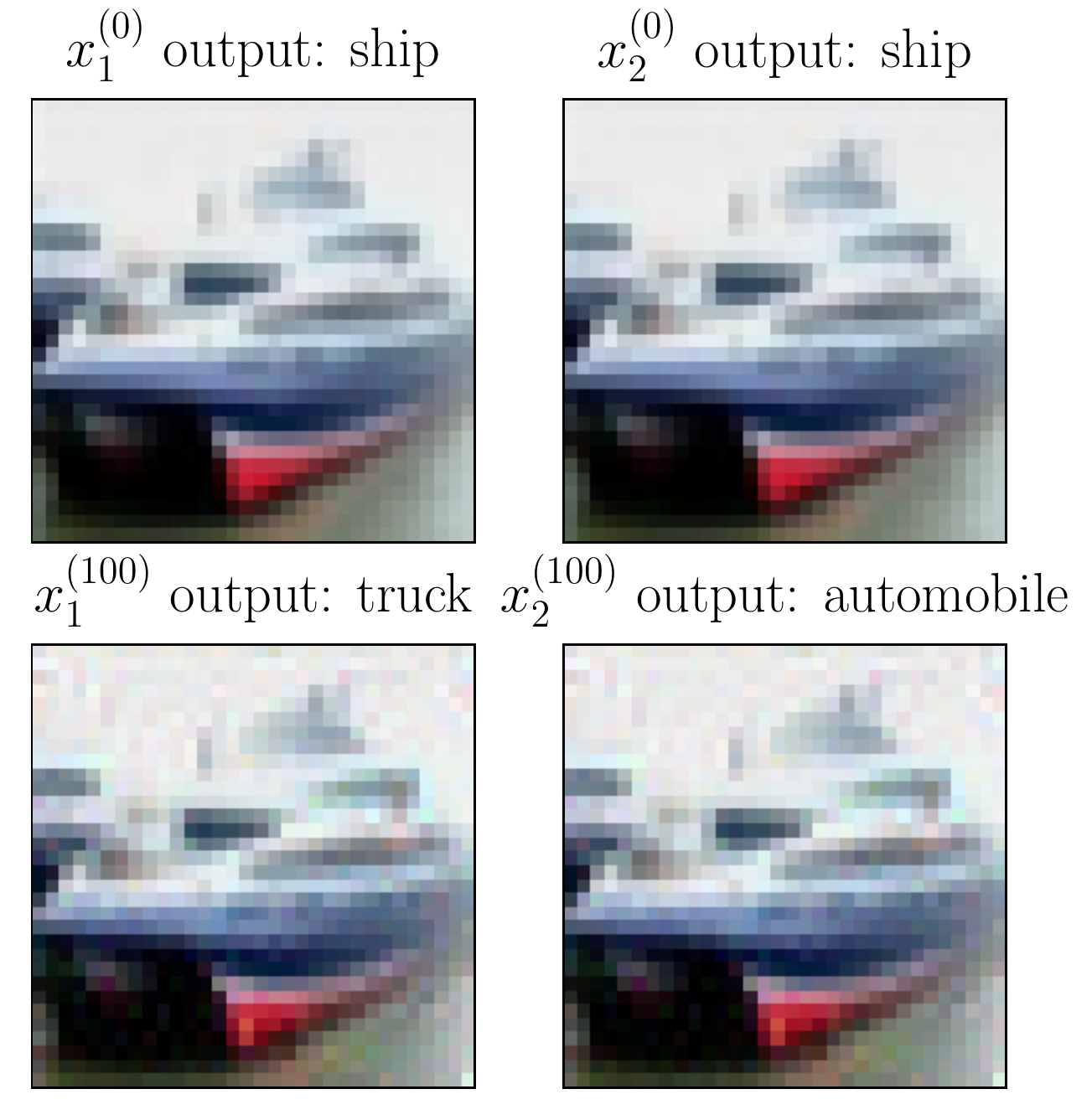}
}
\hfill
\subfloat[\label{sfig::CIFAR10_example:c}]{
\includegraphics[width=0.3\linewidth]{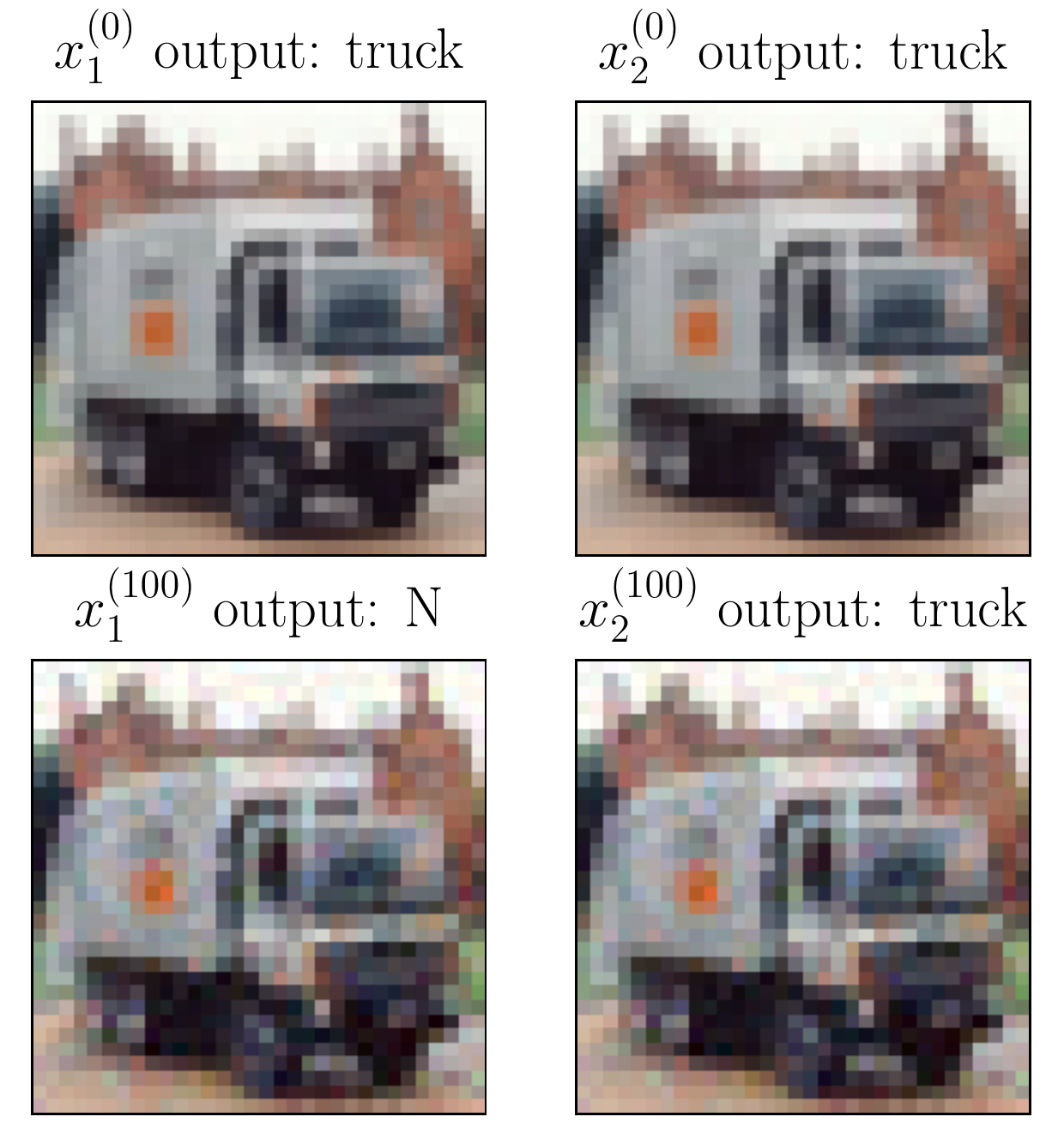}
}\\
\subfloat[\label{sfig::CIFAR10_example:d}]{
\includegraphics[width=0.3\linewidth]{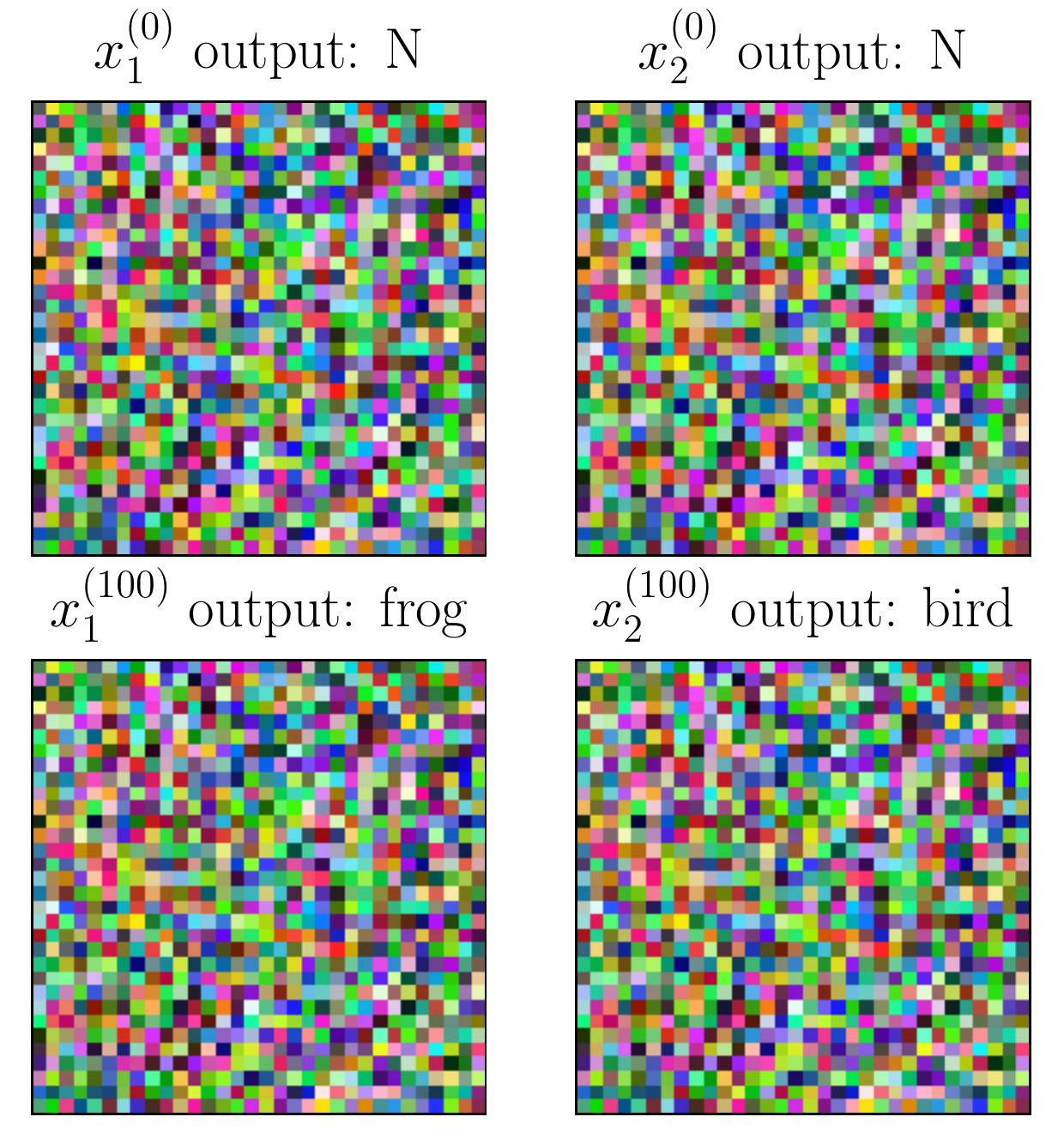}
}
\hfill
\subfloat[\label{sfig::CIFAR10_example:e}]{
\includegraphics[width=0.3\linewidth]{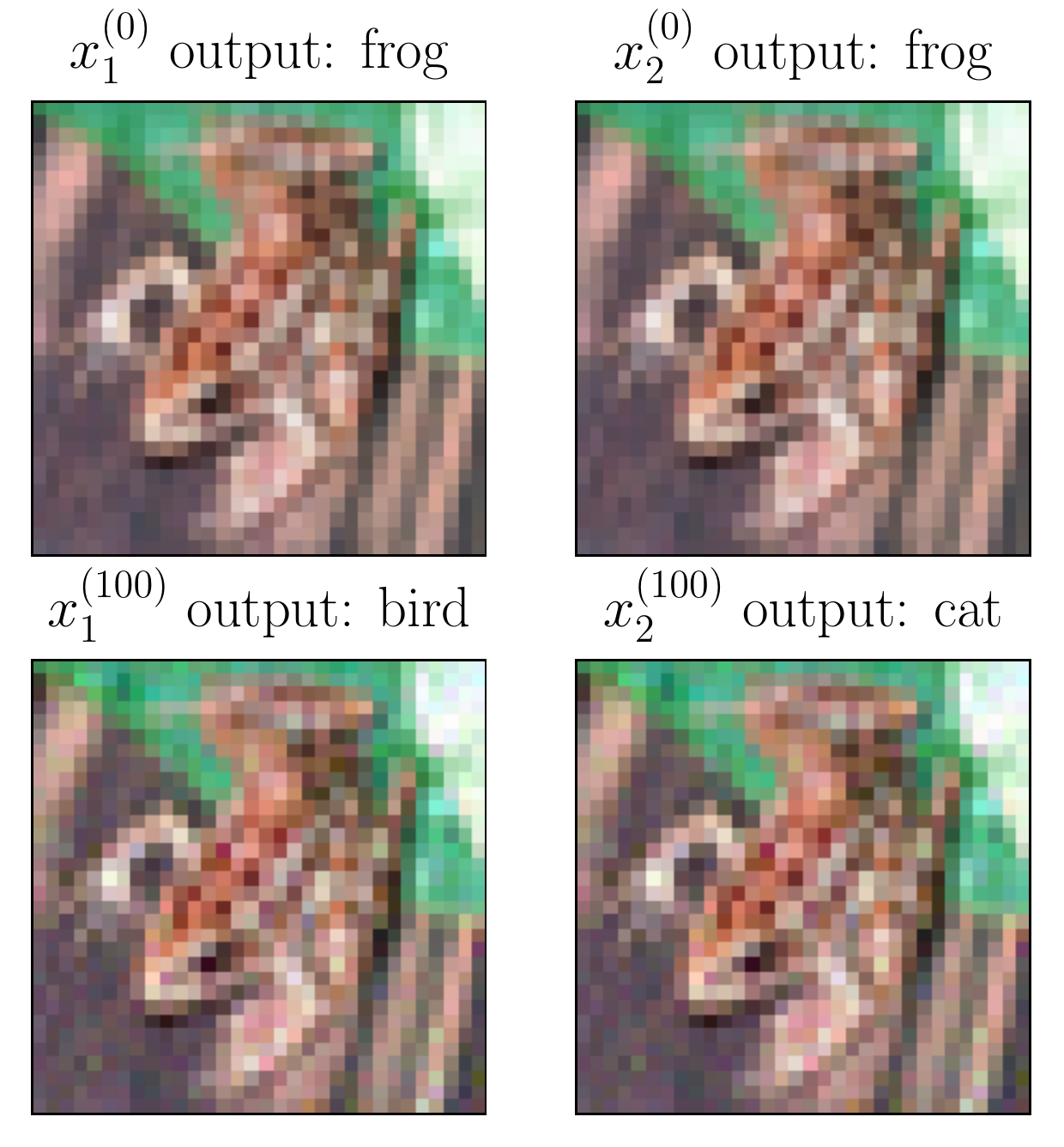}
}
\hfill
\subfloat[\label{sfig::CIFAR10_example:f}]{
\includegraphics[width=0.32\linewidth]{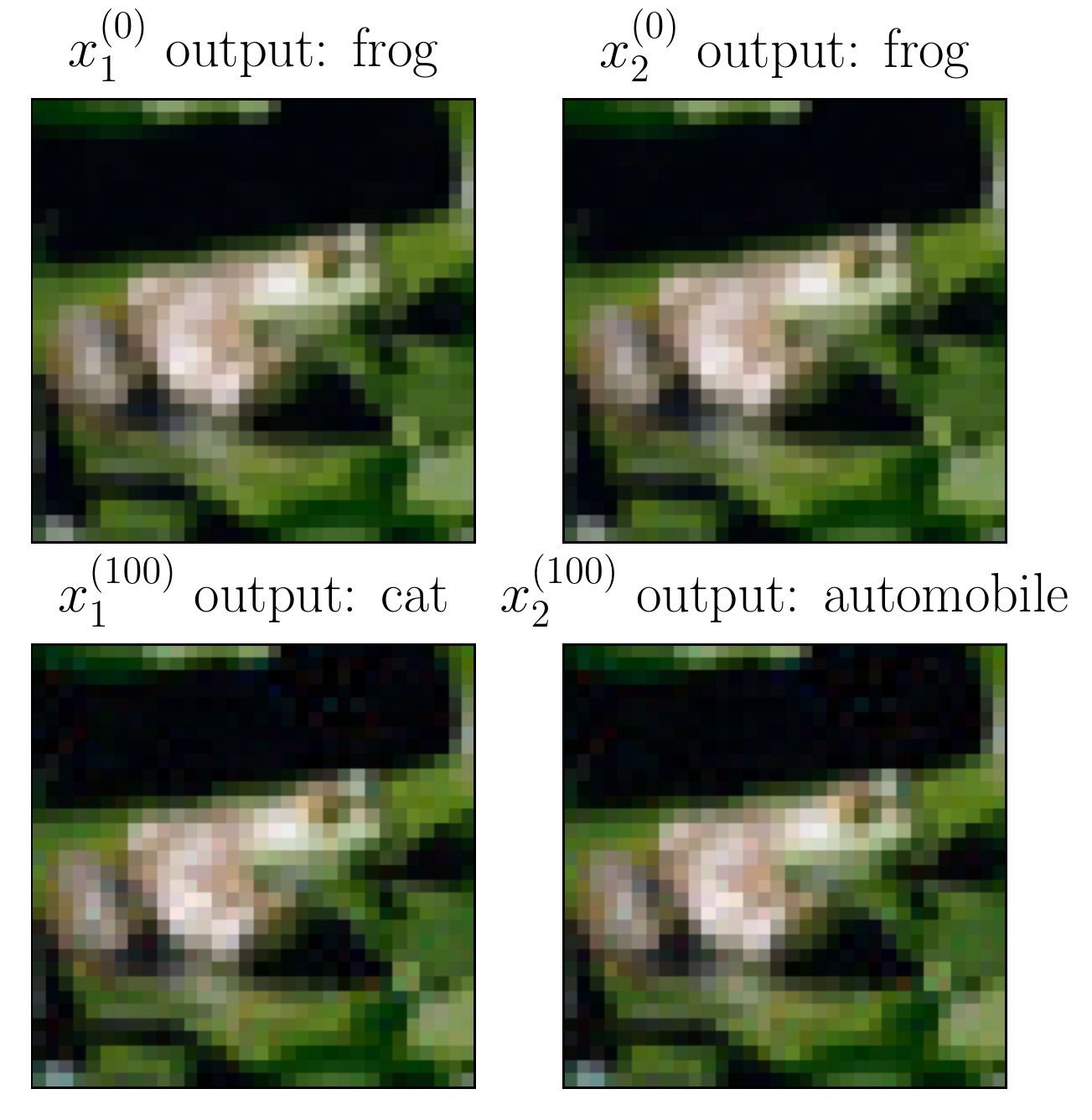}
}
\caption{Global adversarial sample pairs for the CIFAR10 model generated by GEVMCMC.}
\label{fig:CIFAR10_example}
\end{figure}

\begin{figure}
\centering
\subfloat[\label{sfig::MNIST_example:a}]{
\includegraphics[width=0.3\linewidth]{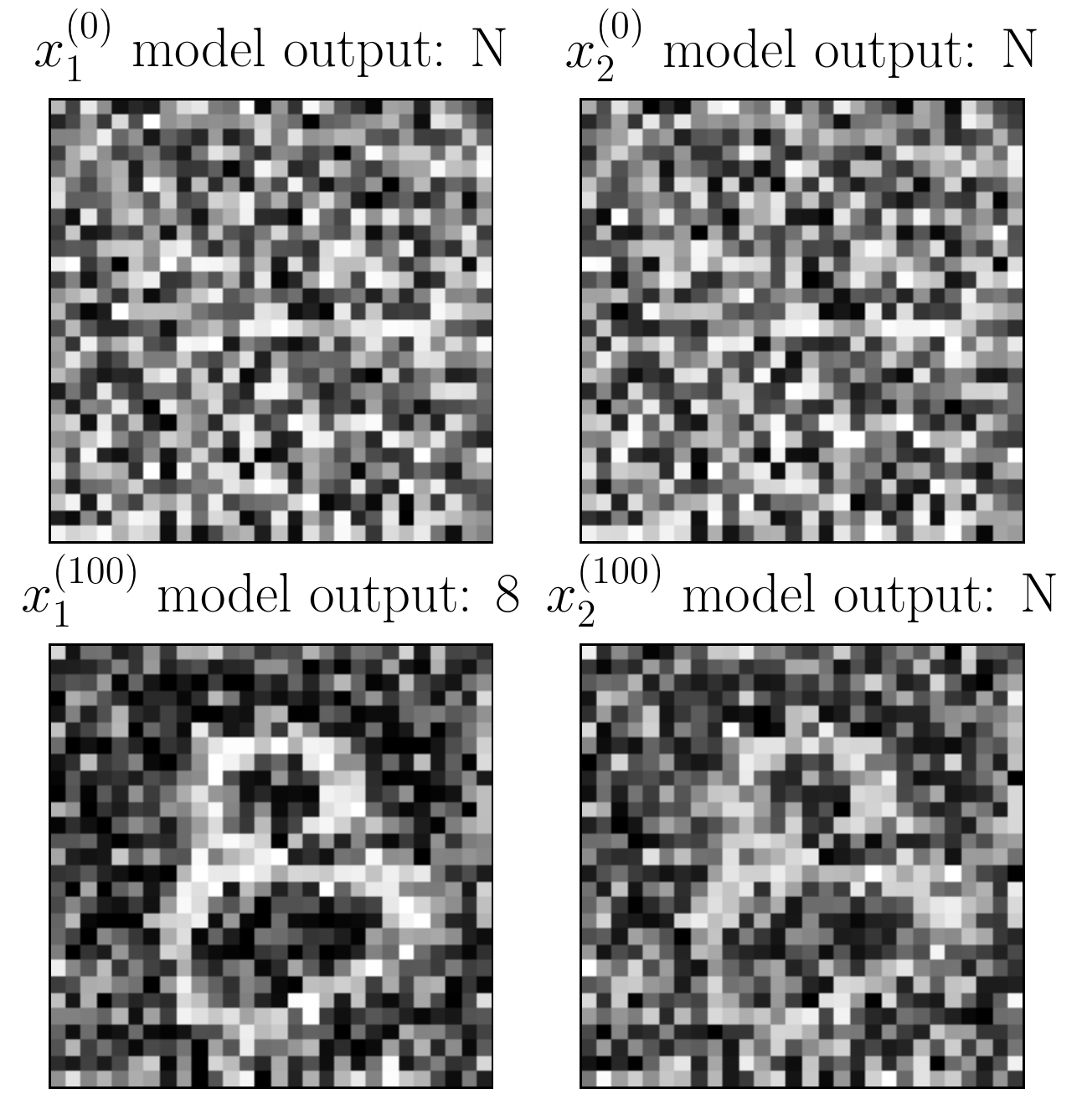}
}
\hfill
\subfloat[\label{sfig::MNIST_example:b}]{
\includegraphics[width=0.3\linewidth]{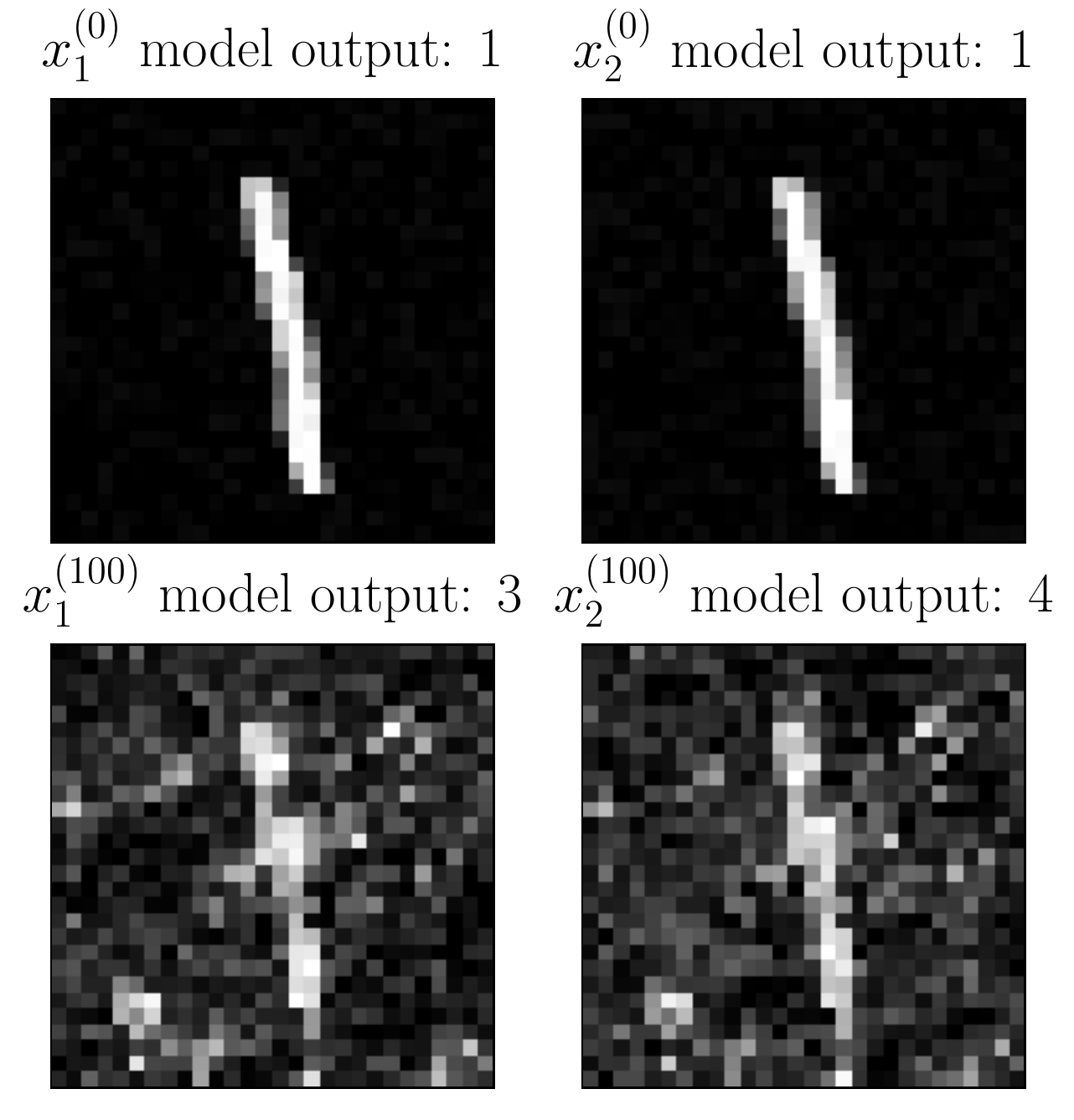}
}
\hfill
\subfloat[\label{sfig::MNIST_example:c}]{
\includegraphics[width=0.3\linewidth]{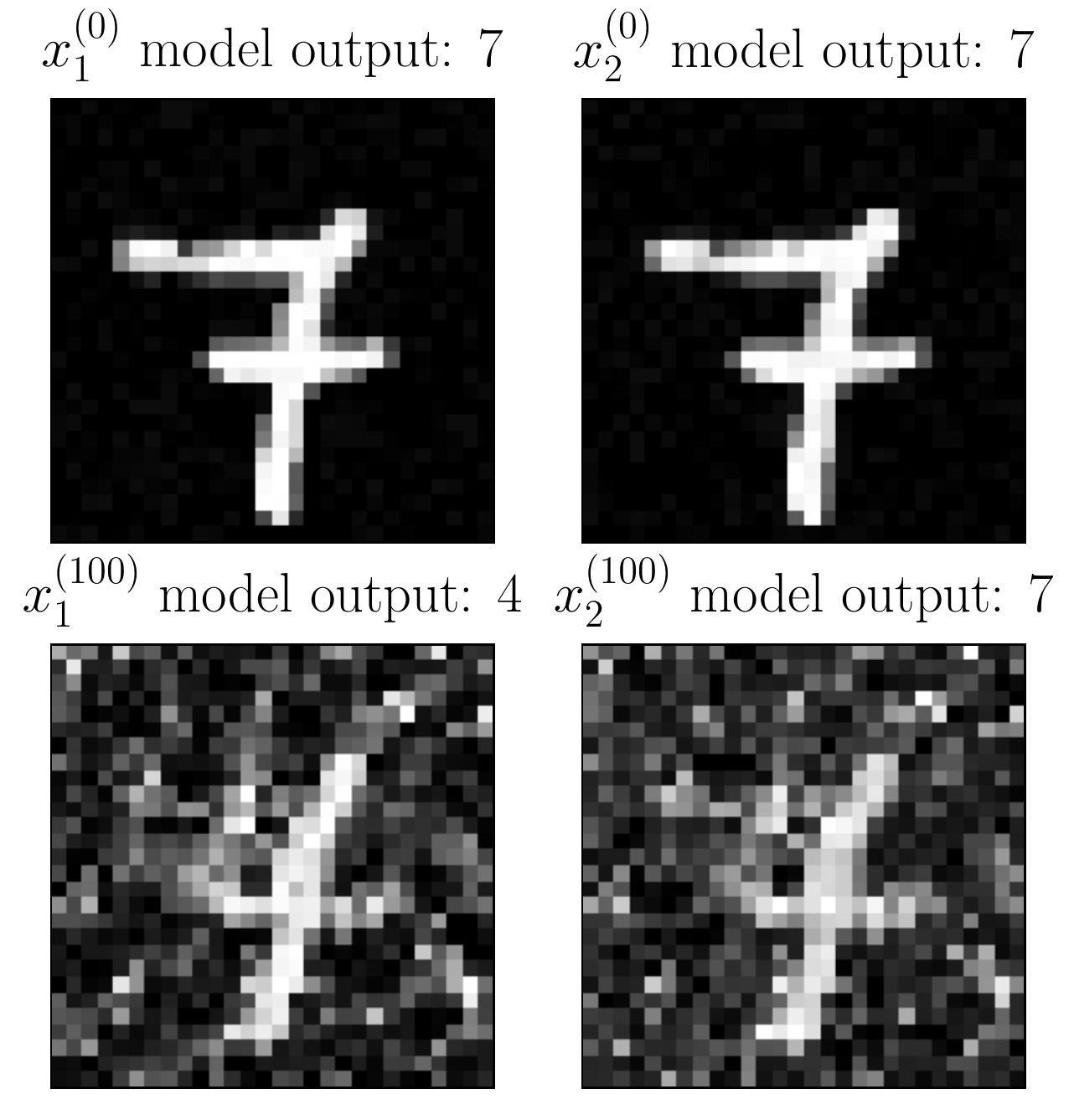}
}\\
\subfloat[\label{sfig::MNIST_example:d}]{
\includegraphics[width=0.3\linewidth]{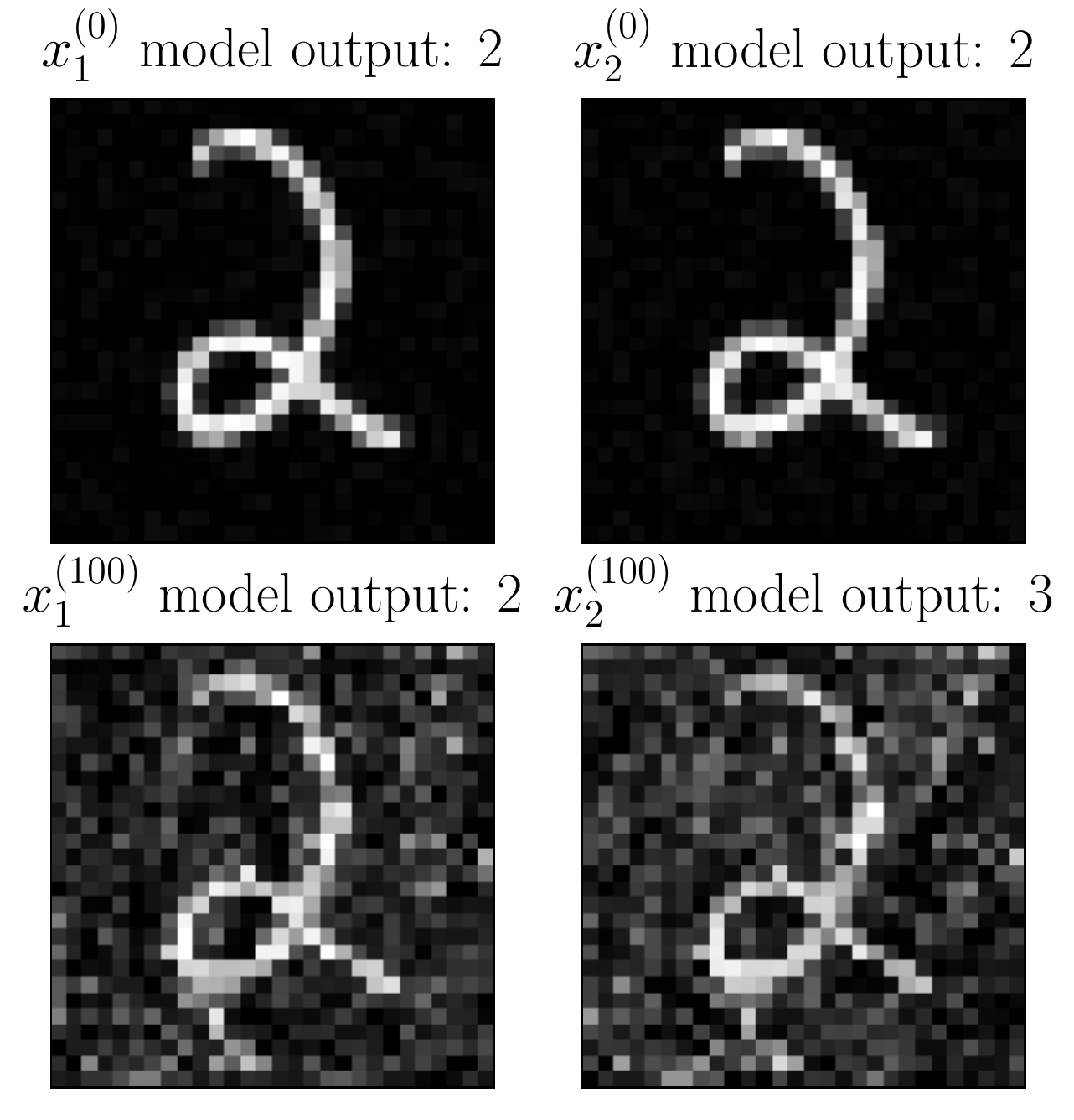}
}
\hfill
\subfloat[\label{sfig::MNIST_example:e}]{
\includegraphics[width=0.3\linewidth]{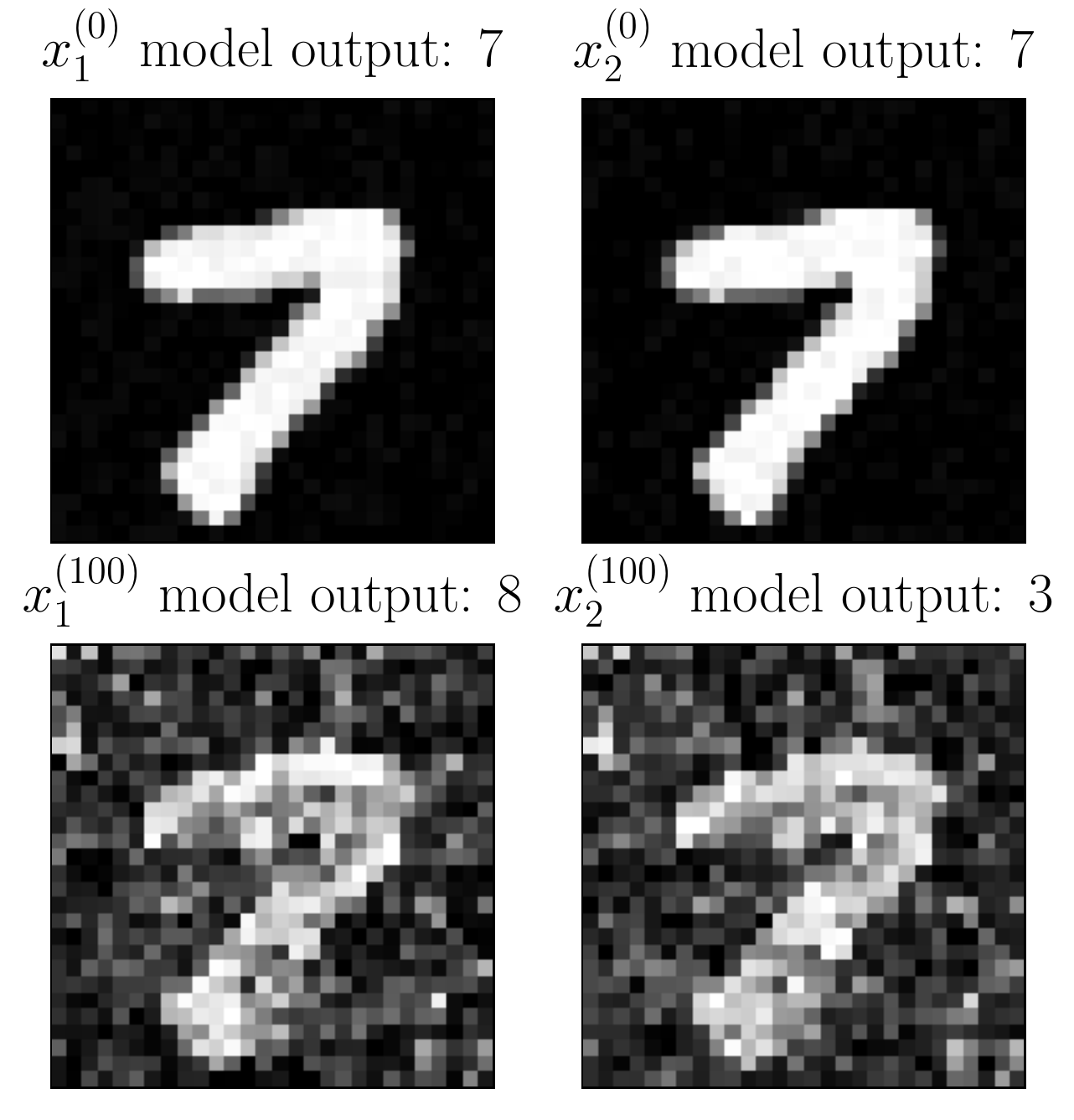}
}
\hfill
\subfloat[\label{sfig::MNIST_example:f}]{
\includegraphics[width=0.3\linewidth]{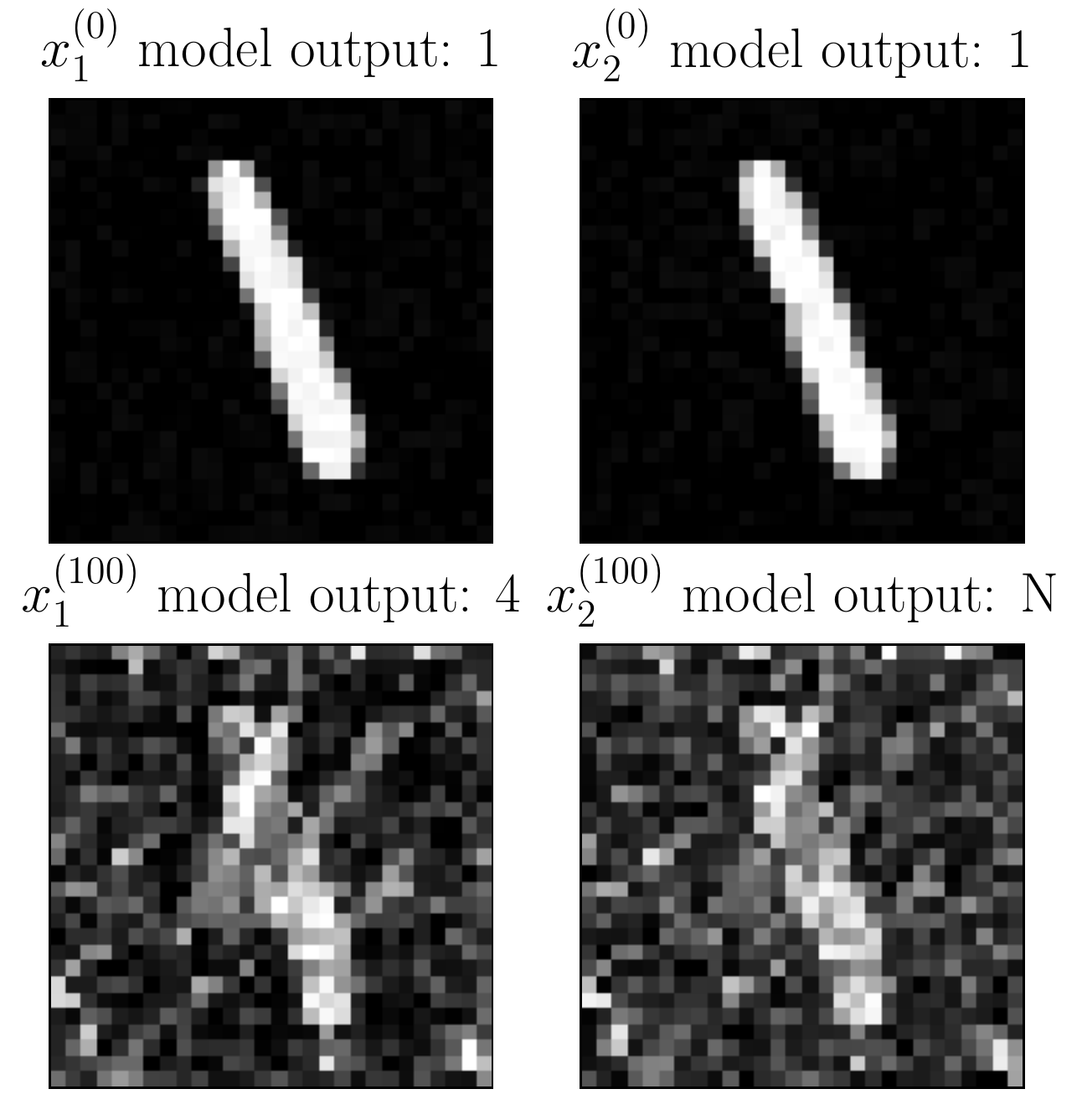}
}
\caption{Global adversarial sample pairs for the MNIST model generated by GEVMCMC.}
\label{fig:MNIST_example}
\end{figure}

\paragraph{Generated global adversarial sample pairs.}

The proposed global adversarial attack methods can generate diverse global adversarial pairs which are rather different from typical local attacks, representing a new type of DNN vulnerability.
Fig. \ref{fig:CIFAR10_example} and Fig. \ref{fig:MNIST_example} show a set of global adversarial example pairs  generated by  GEVMCMC for the two datasets.  In each sub-figure, the two images on the top are the starting pair of two identical starting images. The two at the bottom are the final global adversarial pair generated after 100 rounds of optimization steps. ``N" indicates that the class label predicated by the model is ``meaningless". 

Compared to standard local adversarial attacks, the proposed global adversarial pairs are much more diverse and intriguing. 
For instance, it is possible to start with  two identical random meaningless image but end up with some other two random images that are very similar to each other but have  different legitimate class labels predicted by the model such as ones in Fig. \ref{sfig::CIFAR10_example:a} and \ref{sfig::CIFAR10_example:d}. We can also start from two identical images of a legitimate predicted class, and end up with two images with predicated labels that are different from each other and are also different from the starting label, as shown in Fig. \ref{sfig::CIFAR10_example:b}, \ref{sfig::CIFAR10_example:e}, \ref{sfig::CIFAR10_example:f}, \ref{sfig::MNIST_example:b} and \ref{sfig::MNIST_example:e}. Clearly, the existing local attacks such as FGSM, IFGSM, and PGD are not able to generate such complex global adversarial scenarios, which reveal additional hidden vulnerabilities of the model.  

Importantly, the existing local adversarial attacks cannot explore the input space beyond the training or testing dataset due to the perturbation constraint. In contrast, the proposed global adversarial attacks are very appealing in the following way: they may find a path towards unseen input space and check the model robustness along the way. For instance, we may start from some random testing image (Fig. \ref{sfig::MNIST_example:a}) or original testing image (Fig. \ref{sfig::MNIST_example:c}), and end up with a completely different image pair which may be both recognized by the humans as a legitimate class, however, the model predicts different labels for them. For instance, the final pairs in the Fig.~\ref{sfig::MNIST_example:a} and Fig.~\ref{sfig::MNIST_example:c} may be recognized as ``8" and ``4", respectively, which are completely different from their starting label.


\begin{table}
  \caption{Natural MNIST model (w/o local adversarial training) adversarial attack results.}
  \label{tab:MNIST_nat}
  \centering
  \begin{tabular}{ccccccc}
    \toprule
    \multirow{2}{*}{Method}&\multicolumn{3}{c}{Start from Original Test Image}&\multicolumn{3}{c}{Start from Random Test Image} \\
    \cmidrule(r){2-4}\cmidrule(r){5-7}
         & Attack Rate & Max Loss & Avg. Loss & Attack Rate & Max Loss & Avg. Loss\\
    \midrule
    L-FGSM   & 18.18\% & 20.94 & 0.77 & 5.82\%  & 6.43 & 0.16 \\
    L-IFGSM  & 29.80\% & 21.76 & 1.17 & 51.96\% & 7.64 & 1.18 \\
    L-PGD    & 29.59\% & 21.73 & 1.16 & 49.44\% & 7.57 & 1.13 \\
    G-FGSM   & 99.14\% & 23.40 & 10.78 & 99.01\% & 20.04 & 9.62 \\
    G-IFGSM  & 99.92\% & 22.21 & 9.19  & 100.00\% & 12.05 & 5.94 \\
    G-PGD    & 99.94\% & 20.44 & 10.18 & 100.00\% & 11.94 & 6.58 \\
    GEVMCMC  & 99.93\% & 24.93 & 14.91 & 100.00\% & 19.98 & 12.67 \\
    \bottomrule
  \end{tabular}
\end{table}

\begin{table}
  \caption{Adversarially-trained MNIST model adversarial attack results.}
  \label{tab:MNIST_adv}
  \centering
  \begin{tabular}{ccccccc}
    \toprule
    \multirow{2}{*}{Method}&\multicolumn{3}{c}{Start from Original Test Image}&\multicolumn{3}{c}{Start from Random Test Image} \\
    \cmidrule(r){2-4}\cmidrule(r){5-7}
         & Attack Rate & Max Loss & Avg. Loss & Attack Rate & Max Loss & Avg. Loss\\
    \midrule
    L-FGSM   & 3.28\% & 15.14 & 0.11 & 0.00\% & 0.00 & 0.00 \\
    L-IFGSM  & 3.89\% & 15.56 & 0.13 & 0.00\% & 0.06 & 0.00 \\
    L-PGD    & 3.88\% & 15.57 & 0.13 & 0.00\% & 0.05 & 0.00 \\
    G-FGSM   & 98.07\% & 17.79 & 8.19 & 97.50\% & 19.09 & 7.63 \\
    G-IFGSM  & 99.47\% & 14.25 & 7.84 & 99.56\% & 17.81 & 11.43 \\
    G-PGD    & 99.50\% & 18.08 & 8.61 & 99.58\% & 19.59 & 12.07 \\
    GEVMCMC  & 99.38\% & 18.28 & 9.91 & 99.55\% & 18.06 & 12.04 \\
    \bottomrule
  \end{tabular}
\end{table}

\begin{table}
  \caption{Natural CIFAR10 model (w/o local adversarial training) adversarial attack results.}
  \label{tab:CIFAR10_nat}
  \centering
  \begin{tabular}{ccccccc}
    \toprule
    \multirow{2}{*}{Method}&\multicolumn{3}{c}{Start from Original Test Image}&\multicolumn{3}{c}{Start from Random Test Image} \\
    \cmidrule(r){2-4}\cmidrule(r){5-7}
         & Attack Rate & Max Loss & Avg. Loss & Attack Rate & Max Loss & Avg. Loss\\
    \midrule
    L-FGSM   & 26.77\% & 11.42 & 1.90 & 0.00\% & 0.01 & 0.00 \\
    L-IFGSM  & 44.80\% & 12.06 & 3.58 & 0.00\% & 0.02 & 0.00 \\
    L-PGD    & 43.92\% & 12.04 & 3.51 & 0.00\% & 0.02 & 0.00 \\
    G-FGSM   & 95.96\% & 12.75 & 8.72  & 95.15\% & 11.10 & 5.48 \\
    G-IFGSM  & 99.68\% & 12.90 & 11.02 & 98.36\% & 10.76 & 6.45 \\
    G-PGD    & 99.71\% & 13.55 & 11.30 & 98.33\% & 11.02 & 6.70 \\
    GEVMCMC  & 99.58\% & 13.18 & 8.94  & 98.37\% & 11.02 & 7.18 \\
    \bottomrule
  \end{tabular}
\end{table}

\begin{table}
  \caption{Adversarially-trained CIFAR10 model adversarial attack results.}
  \label{tab:CIFAR10_adv}
  \centering
  \begin{tabular}{ccccccc}
    \toprule
    \multirow{2}{*}{Method}&\multicolumn{3}{c}{Start from Original Test Image}&\multicolumn{3}{c}{Start from Random Test Image} \\
    \cmidrule(r){2-4}\cmidrule(r){5-7}
         & Attack Rate & Max Loss & Avg. Loss & Attack Rate & Max Loss & Avg. Loss\\
    \midrule
    L-FGSM   & 19.29\% & 10.97 & 0.83 & 0.00\% & 0.00 & 0.00 \\
    L-IFGSM  & 21.25\% & 11.03 & 0.92 & 0.00\% & 0.00 & 0.00 \\
    L-PGD    & 21.19\% & 11.03 & 0.92 & 0.00\% & 0.00 & 0.00 \\
    G-FGSM   & 95.29\% & 9.62  & 4.11 & 90.47\% & 6.57 & 2.52 \\
    G-IFGSM  & 98.23\% & 9.83  & 4.60 & 95.85\% & 6.99 & 2.94 \\
    G-PGD    & 98.26\% & 10.54 & 4.77 & 95.88\% & 6.97 & 3.03 \\
    GEVMCMC  & 98.25\% & 10.89 & 5.47 & 95.89\% & 6.66 & 3.67 \\
    \bottomrule
  \end{tabular}
\end{table}

\paragraph{Local vs. global adversarial attacks}

Table \ref{tab:MNIST_nat}-\ref{tab:CIFAR10_adv} show the global adversarial attack results for the MNIST and CIFAR10 models without and with local adversarial training. If the model predictions for the two examples in a generated global adversarial pair are different, we regard this pair as one successful global attack. In these tables, the attack rate  is defined as the ratio between the successful number of attacks over the total number of trails which is 10,000. The Max loss and Avg. loss are the maximum loss and average loss found in the attack process. 

Attacking the natural MNIST and CIFAR10 models using a local attack method can reach a reasonablly high attack rate. For instance, L-PGD has an attack rate of $29.59\%$ in the case of the natural MNIST, which drops down to $3.88\%$ for the case of the adversarially-trained MNIST model, showing that adversarial training using local adversarial examples does improve the defense to such local attacks. 
However, all global adversarial attack methods achieve almost 100.00\% attack rate and produce much higher average loss values compared to the local attack methods, regardless whether local adversarial training is performed or not. It is evident that adversarial training based on local adversarial examples shows none or little defense to global attacks. This indicates the effectiveness of the proposed global attack methods, and equally importantly, suggests that global adversarial examples defined in this paper must be coped with when training robust DNN models.

\begin{table}
  \caption{Comparison between GEVMCMC and other proposed global attacks when starting from the same 100 original test (``Test") or 100 random testing (``Rand") example pairs. Each entry shows the number of cases out of the total 100 cases where the final adversarial pair generated by GEVMCMC has a loss higher than the one generated by the other method.   
  }
  \label{tab:cnt}
  \centering
  \begin{tabular}{ccccccccc}
    \toprule
    \multirow{3}{*}{Method}&\multicolumn{4}{c}{MNIST}&\multicolumn{4}{c}{CIFAR10} \\
    \cmidrule(r){2-5}\cmidrule(r){6-9}
        & \multicolumn{2}{c}{Natural}&\multicolumn{2}{c}{Adv.-Trained}&\multicolumn{2}{c}{Natural}&\multicolumn{2}{c}{Adv.-Trained}\\
        \cmidrule(r){2-3}\cmidrule(r){4-5}\cmidrule(r){6-7}\cmidrule(r){8-9}
        & Test & Rand & Test & Rand & Test & Rand & Test & Rand \\
    \midrule
    G-FGSM   & 86 & 95  & 69 & 81 & 33 & 72 & 76 & 67 \\
    G-IFGSM  & 98 & 100 & 92 & 90 & 69 & 85 & 84 & 92 \\
    G-PGD    & 97 & 100 & 80 & 67 & 12 & 73 & 74 & 90 \\
    \bottomrule
  \end{tabular}
\end{table}

\begin{figure}
\subfloat[Testing image starting image pair\label{sfig:MNIST_nat_test}]{
\includegraphics[width=0.48\linewidth]{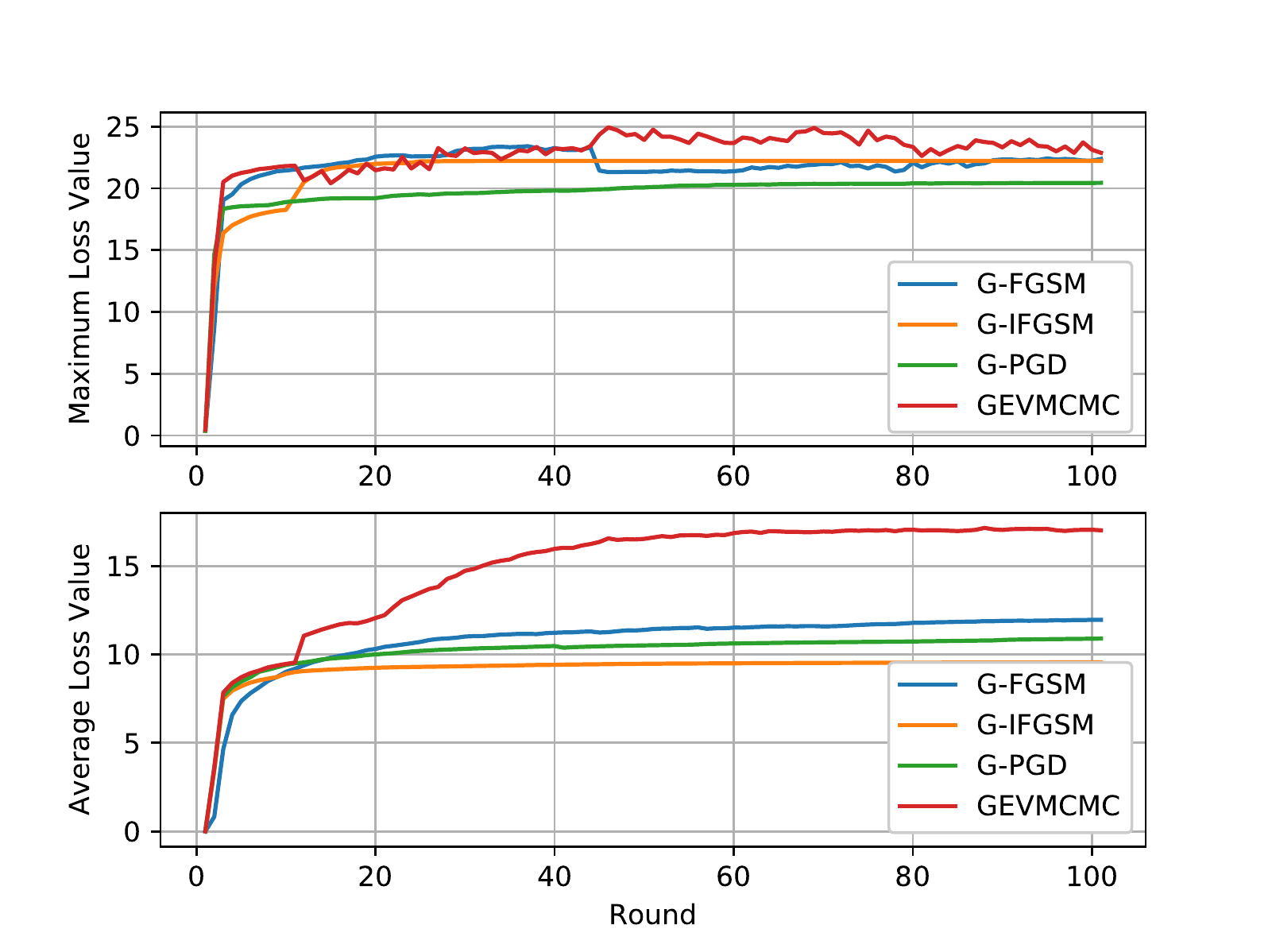}
}
\hfill
\subfloat[Random image starting image pair\label{sfig:MNIST_nat_rand}]{
\includegraphics[width=0.48\linewidth]{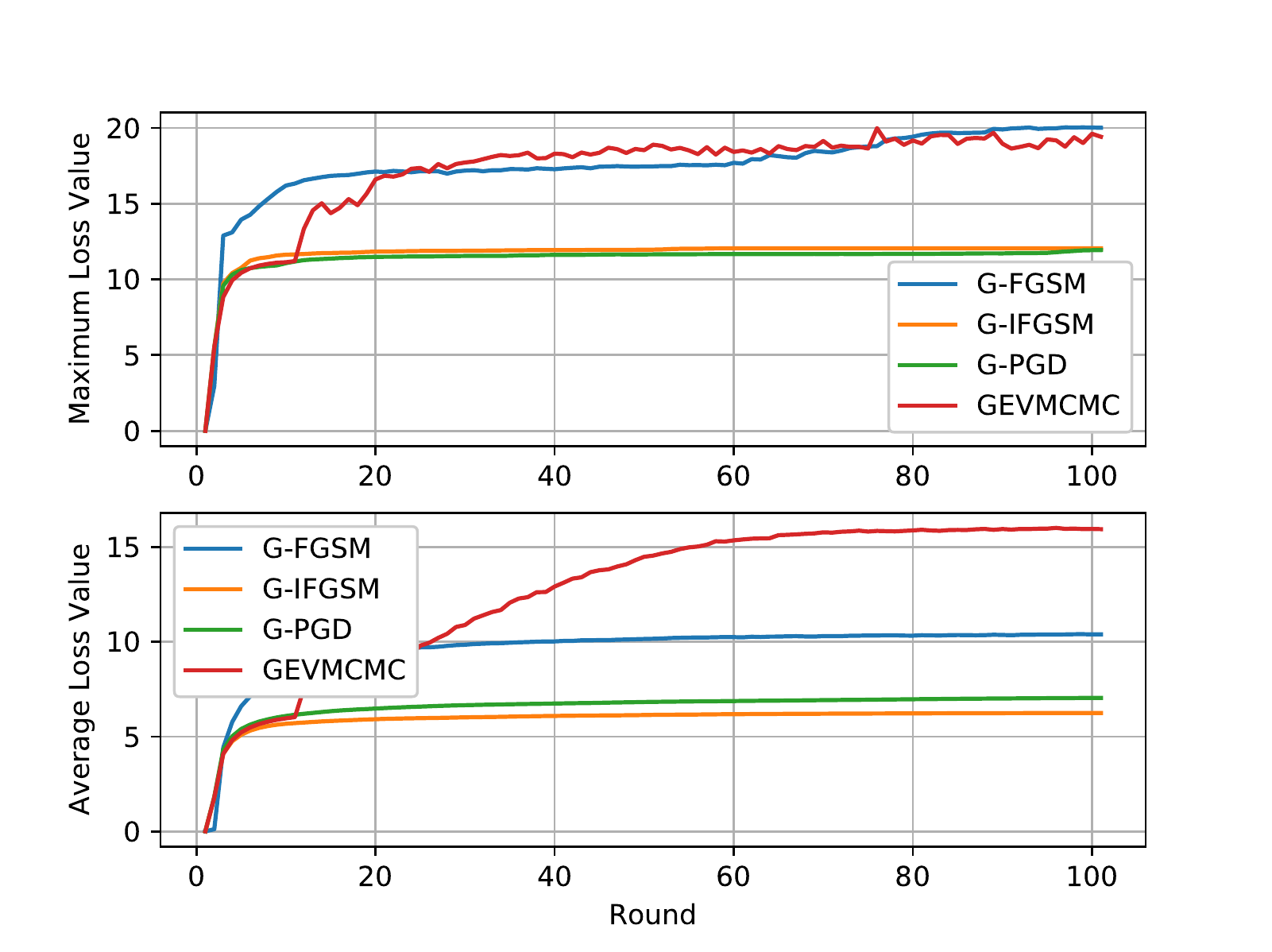}
}
\caption{Loss value found by global adversarial attack in each round for the natural MNIST model.}
\label{fig:MNIST_nat}
\end{figure}

\begin{figure}
\subfloat[Testing image starting image pair\label{sfig:MNIST_adv_test}]{
\includegraphics[width=0.48\linewidth]{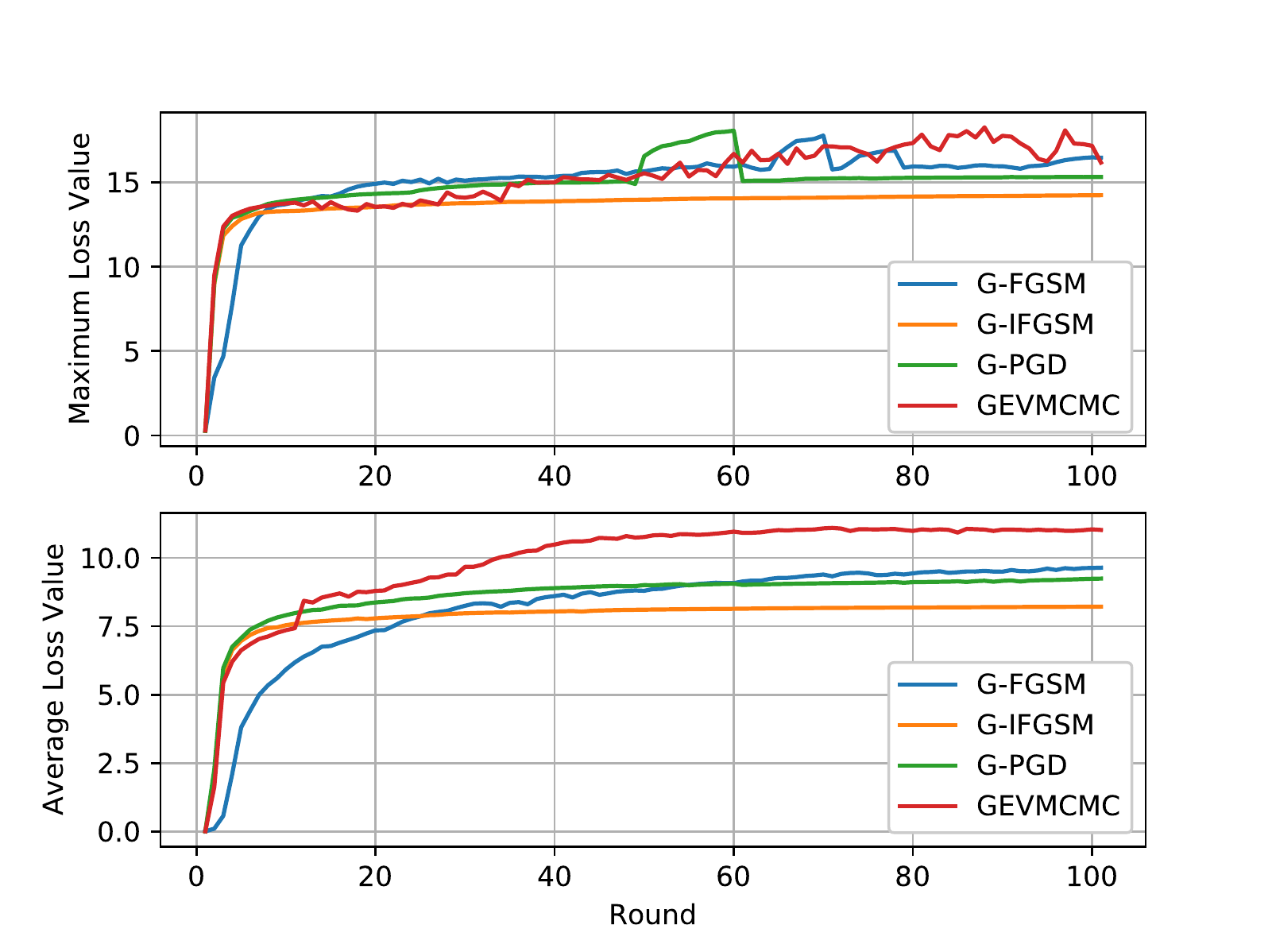}
}
\hfill
\subfloat[Random image starting image pair\label{sfig:MNIST_adv_rand}]{
\includegraphics[width=0.48\linewidth]{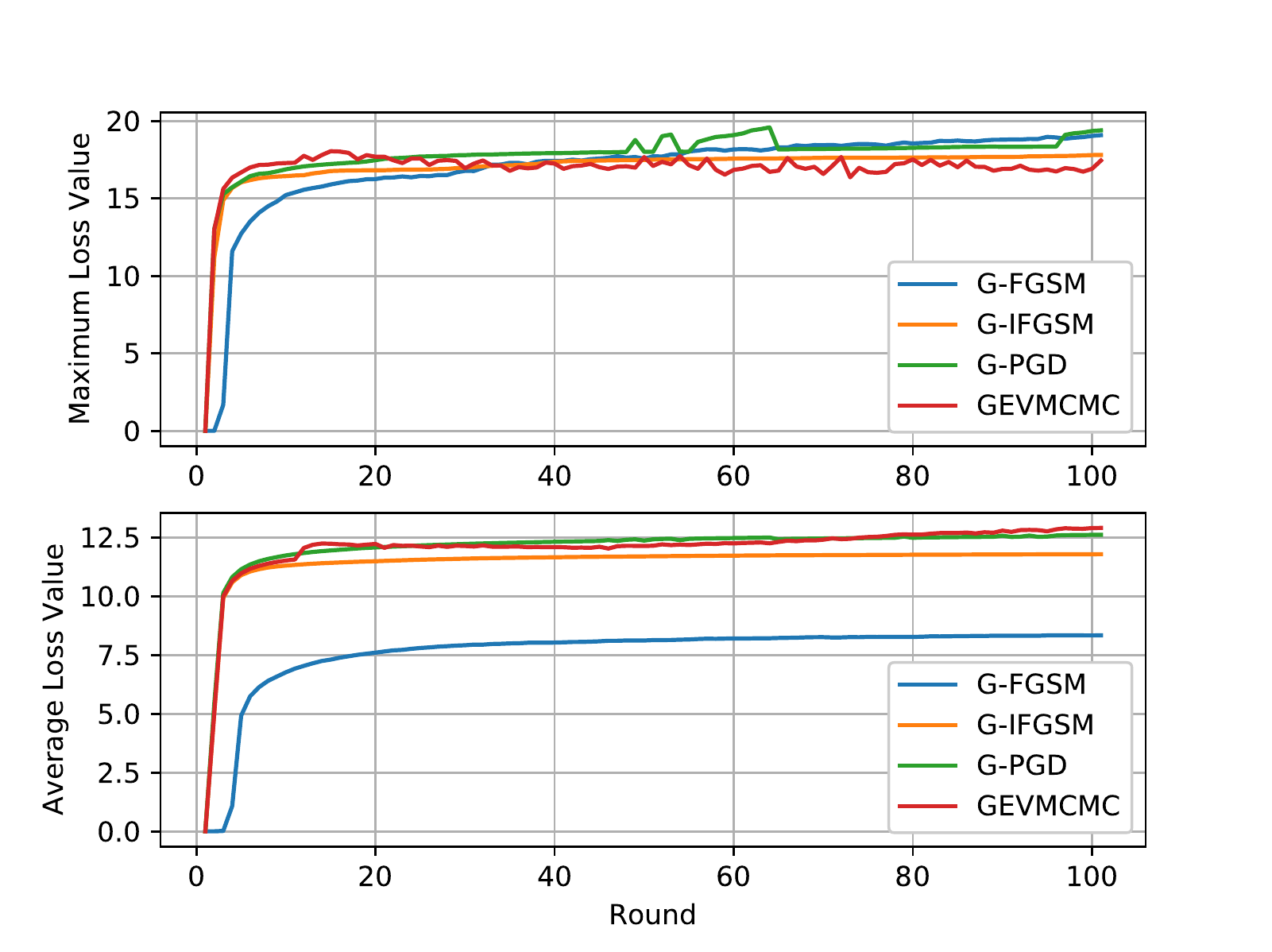}
}
\caption{Loss value found by global adversarial attack in each round for the adversarially-trained MNIST model.}
\label{fig:MNIST_adv}
\end{figure}

\begin{figure}
\subfloat[Testing image starting image pair\label{sfig:CIFAR10_nat_test}]{
\includegraphics[width=0.48\linewidth]{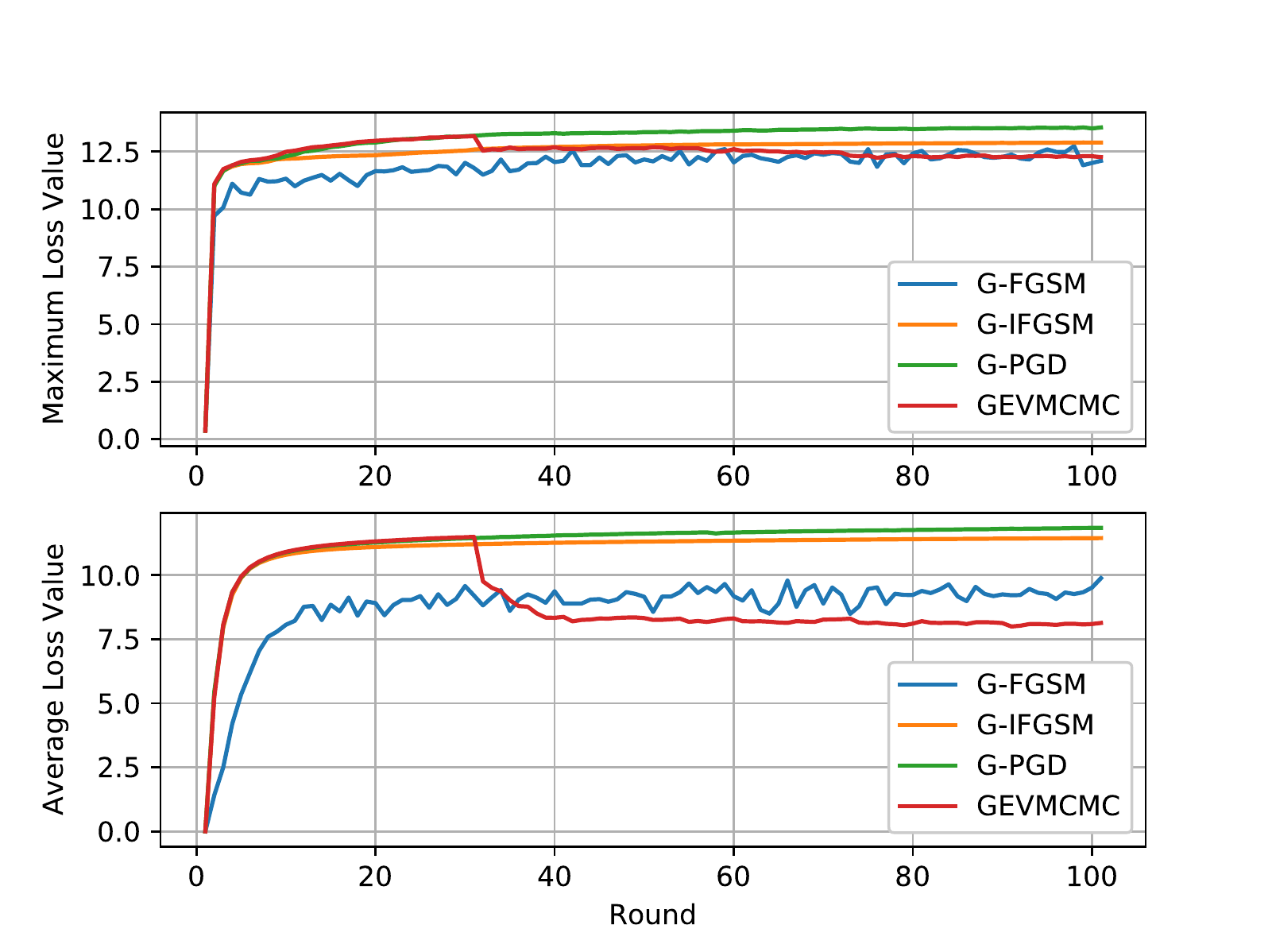}
}
\hfill
\subfloat[Random image starting image pair\label{sfig:CIFAR10_nat_rand}]{
\includegraphics[width=0.48\linewidth]{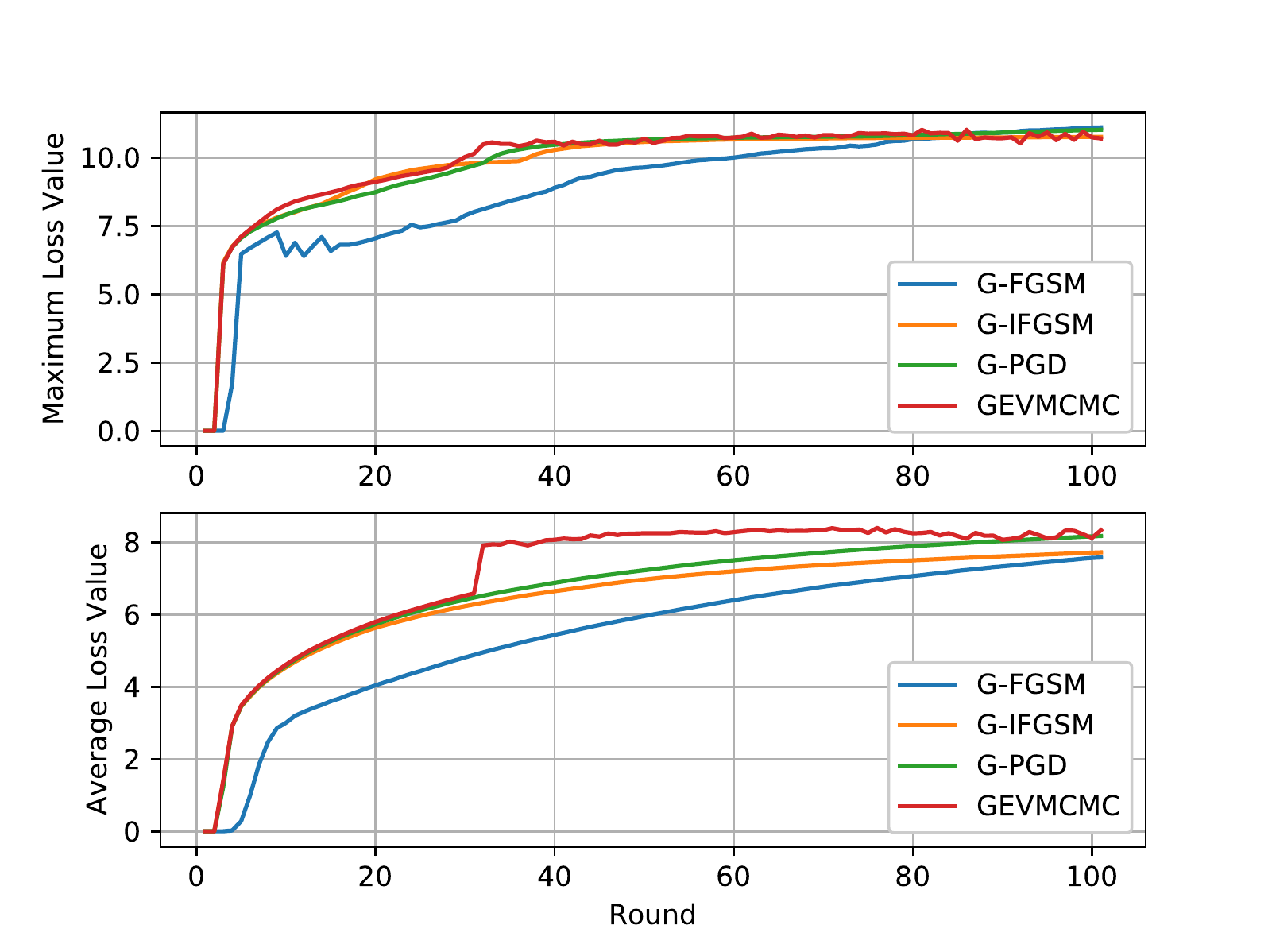}
}
\caption{Loss value found by global adversarial attack in each round for the natural CIFAR10 model.}
\label{fig:CIFAR10_nat}
\end{figure}

\begin{figure}
\subfloat[Testing image starting image pair\label{sfig:CIFAR10_adv_test}]{
\includegraphics[width=0.48\linewidth]{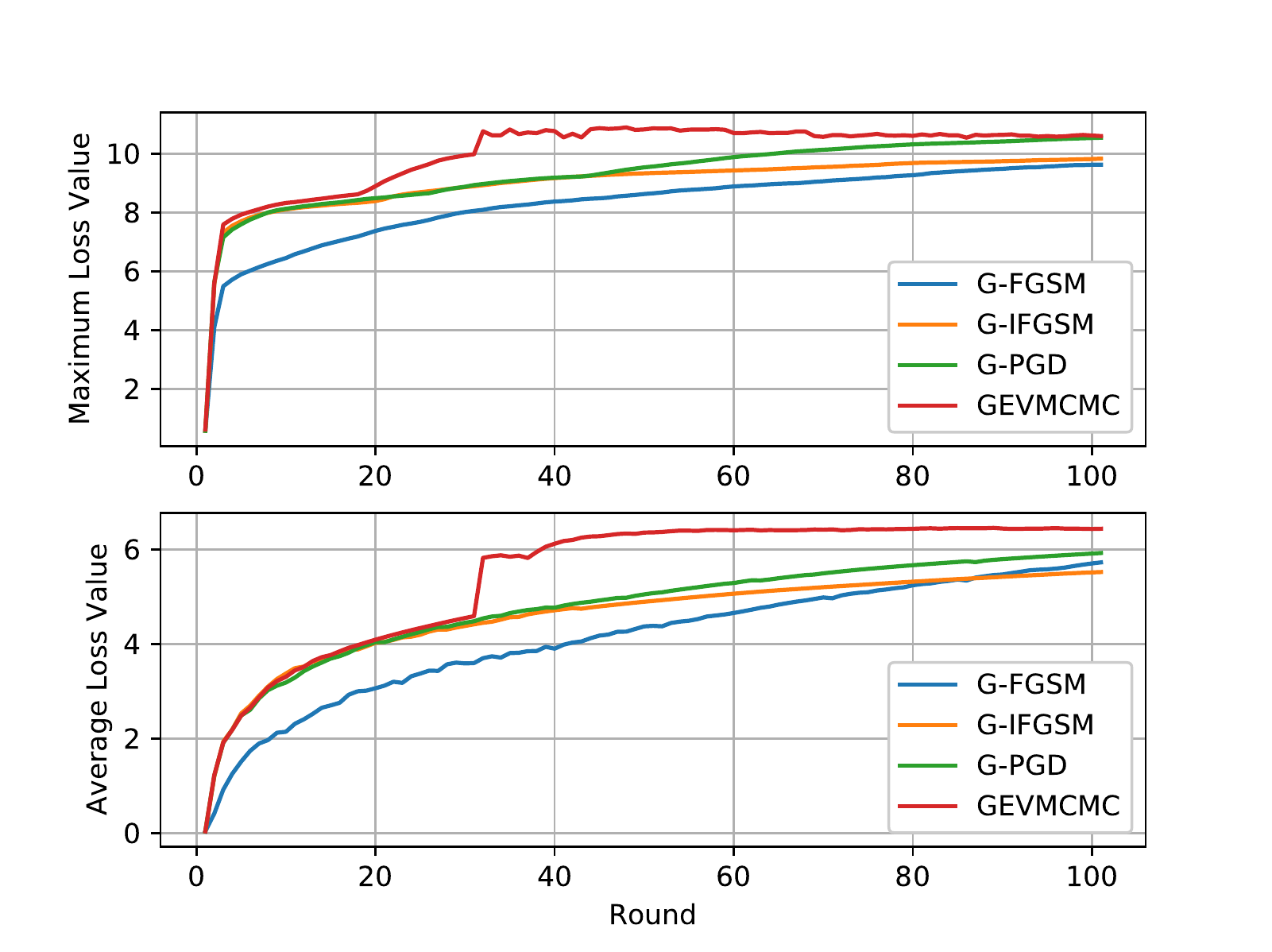}
}
\hfill
\subfloat[Random image starting image pair\label{sfig:CIFAR10_adv_rand}]{
\includegraphics[width=0.48\linewidth]{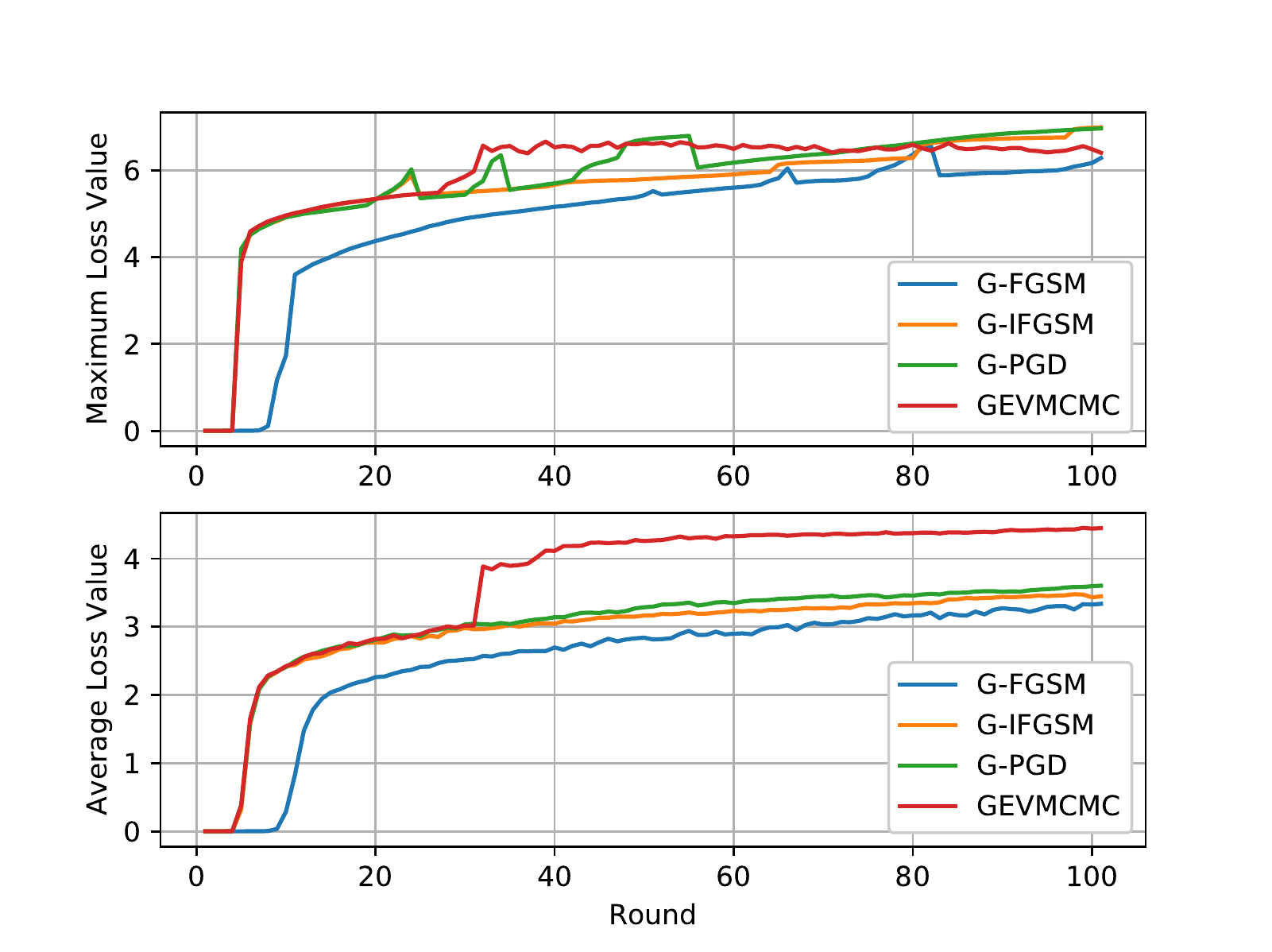}
}
\caption{Loss value found by global adversarial attack in each round for the adversarially-trained CIFAR10 model.}
\label{fig:CIFAR10_adv}
\end{figure}

\paragraph{Comparison of the proposed global adversarial attack methods}

Table \ref{tab:cnt} compares the two  types of the proposed global attack methods when starting from the same 100 original test (``Test") or 100 random testing (``Rand") example pairs.
Each entry shows the number of cases out of the total 100 cases where the final adversarial pair generated by GEVMCMC has a loss higher than the one generated by the other method.  Most entries in the table are larger than 50, implying that  GEVMCMC finds worse adversarial example pairs than other global adversarial attack methods. We further show the maximum loss value and average loss value of the adversarial example pairs in each round in Fig. \ref{fig:MNIST_nat} - \ref{fig:CIFAR10_adv}. After the initial warm-up rounds using G-PGD, the loss found by GEVMCMC increases rapidly, and ends up with a much larger value  compared to that of the other global adversarial attack methods, which tend to converge at a local maximum. The only case in which GEVMCMC does not beat other global adversarial attack methods is the attacking of the natural CIFAR10 model starting with the original testing images, showing the overall better effectiveness of GEVMCMC.  

\section{Conclusion}
We propose a new global adversarial example pair concept and formulate the corresponding global adversarial attack problem to assess the robustness of DNNs over the entire input space without human data labeling. We further propose two families of global adversarial attack methods: (\textbf{1}) alternating gradient global adversarial attacks and (\textbf{2}) extreme-value-guided MCMC sampling global attack (GEVMCMC), demonstrating that DNN models even trained with local adversarial training are vulnerable to this new type of global attacks.  Our  attack methods are able to generate diverse and intriguing global adversarial which are very different from typical local attacks and shall be taken into consideration when training a robust model. GEVMCMC demonstrates the overall best performance among all proposed global attack methods due to its probabilistic nature. 

\if@0
\subsubsection*{Acknowledgments}
This material is based upon work supported by the Semiconductor Research Corporation (SRC) under Task 2810.024. 
The authors would like to thank High Performance Research Computing (HPRC) at Texas A\&M University for providing computing support. Any opinions, findings, conclusions or recommendations expressed in this material are those of the authors and do not necessarily reflect the views of SRC, Texas A\&M University, and their contractors.
\fi

\bibliographystyle{unsrtnat}
\bibliography{gadv}

\end{document}